\documentclass[letterpaper]{article}
\usepackage{aaai}

\usepackage{comment}
\usepackage{hyperref}
\usepackage{graphicx}%
\usepackage{multirow}%
\usepackage{amsmath,amssymb,amsfonts}%
\usepackage{amsthm}%
\usepackage{mathrsfs}%
\usepackage[title]{appendix}%
\usepackage{xcolor}%
\usepackage{textcomp}%
\usepackage{manyfoot}%
\usepackage{booktabs}%
\usepackage{algorithm}%
\usepackage{algorithmicx}%
\usepackage{algpseudocode}%
\usepackage{listings}%

\usepackage{dingbat}  

\usepackage{pifont}
\usepackage[table]{xcolor}
\usepackage{comment}
\newcommand{\cmark}{\textcolor{green!70!black}{\ding{51}}} 
\newcommand{\xmark}{\textcolor{red}{\ding{55}}} 

\usepackage{svg}

\usepackage{enumitem}
\usepackage{adjustbox}

\usepackage{soul} 

\definecolor{lightblue}{HTML}{9DC2F2}
\definecolor{lightpurple}{HTML}{AB9CCE}

\usepackage{todonotes}
\usepackage{caption}
\usepackage{subcaption}

\newcommand{\best}[1]{\cellcolor{green!25}{$\blacktriangle$~#1}}
\newcommand{\second}[1]{\cellcolor{blue!25}{$\bullet$~#1}}
\newcommand{\third}[1]{\cellcolor{red!25}{$\blacklozenge$~#1}}
\usepackage{eso-pic}

\usepackage{times}
\usepackage{helvet}
\usepackage{courier}
\begin{document}
%


\title{OntoLearner: A Modular Python Library for Ontology Learning with Large Language Models}
\author{
Hamed Babaei Giglou$^{1,}$\thanks{These authors contributed equally to this work.},
Jennifer D'Souza$^{1,}$\footnotemark[1],
Andrei Aioanei$^{1}$,
Nandana Mihindukulasooriya$^{3}$,
S\"{o}ren Auer$^{1,2}$\\[1ex]
$^{1}$TIB -- Leibniz Information Centre for Science and Technology, Hannover, Germany\\
$^{2}$L3S Research Center, Leibniz University of Hannover, Hannover, Germany\\
\texttt{\{hamed.babaei,jennifer.dsouza,andrei.aioanei,auer\}@tib.eu}\\[1ex]
$^{3}$IBM Research, New York, USA\\
\texttt{nandana@ibm.com}\\
}


\AddToShipoutPictureFG{%
    \AtPageUpperLeft{%
        \hspace*{0.82\paperwidth}%
        \raisebox{-1.0cm}{%
            \includegraphics[width=0.16\textwidth]{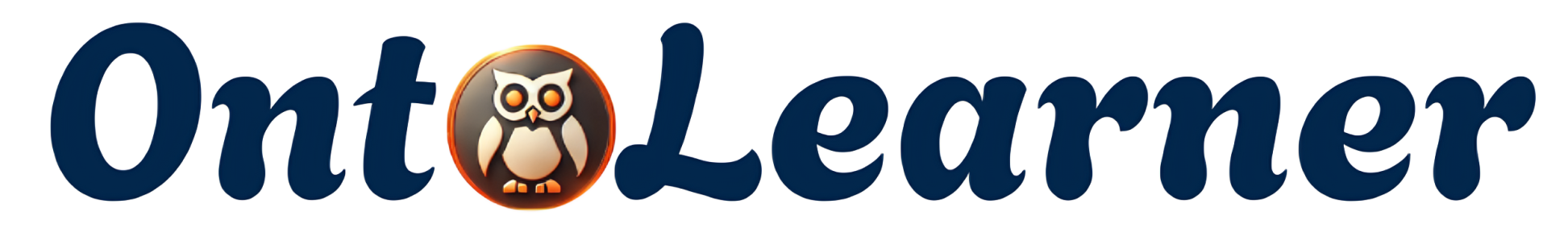}%
        }%
    }%
}

\maketitle

\begin{abstract}
\begin{quote}
Ontology learning (OL) aims to automatically construct structured knowledge models from text, yet progress remains fragmented across methods, domains, and evaluation practices. Despite decades of research, OL lacks a shared infrastructure for systematic evaluation and ontology access. This absence has hindered progress and fragmented research, leaving the central challenges of OL largely unaddressed. We introduce OntoLearner, a modular, cross-domain, and first-of-its-kind framework that unifies ontology access, large language model (LLM)-driven learning pipelines, and standardized benchmarking. OntoLearner releases 180 machine-readable ontologies spanning 22 domains and provides pipeline-ready datasets with train/dev/test splits for three core OL tasks: term typing, taxonomy discovery, and non-taxonomic relation extraction.  Using this infrastructure, we conduct a large-scale empirical study of OL, evaluating 22 retrieval models and 12 LLMs across domains and tasks. The results converge on a finding that reframes the central challenge of OL: failure modes scale with ontological complexity rather than model size or architectural sophistication. The primary bottleneck is not model capability, but a structural mismatch between how models encode knowledge and how ontologies organize it. These findings establish that effective OL is reachable through the cross-domain, multi-task benchmarking enabled by OntoLearner. OntoLearner is open-source (MIT license) at \url{https://github.com/sciknoworg/OntoLearner/}.
\end{quote}
\end{abstract}

\section{Introduction} 
Ontology learning (OL)~\cite{maedche2002ontology} sits at the intersection of symbolic and statistical artificial intelligence (AI), addressing a long-standing challenge in the field: how to combine structured, interpretable knowledge with the flexibility and scalability of data-driven models. Ontologies provide explicit, formal representations of concepts and their relationships, forming a foundation for reasoning, explainability, and interoperability in complex AI systems~\cite{STUDER1998161}.  In emerging neuro-symbolic paradigms~\cite{hitzler2024neuro,bhuyan2024neuro}, such structured knowledge plays a critical role in grounding learning, constraining generation, and enabling more robust generalization beyond surface-level correlations.  However, constructing and maintaining high-quality ontologies remains a bottleneck, limiting their integration into modern AI pipelines.

An ontology engineer may need to determine whether ``cell death'' should be classified as a ``biological process'', a ``pathological state'', or a ``causal event''. Although seemingly narrow, such a decision propagates across every system that relies on the ontology: databases that query it, models that reason over it, and scientists who reuse it. Across thousands of similar decisions spanning hundreds of domains—expanding faster than expert teams can manage—this situation illustrates the central challenge of OL: transforming heterogeneous natural language into structured, machine-interpretable conceptual models that enable reasoning, interoperability, and scientific discovery.

Over the past three decades, numerous approaches have attempted to automate this challenge. The first wave brought lexico-syntactic patterns~\cite{hearst1992automatic} and modular pipelines: Text2Onto~\cite{cimiano2005text2onto} combined probabilistic modeling with linguistic heuristics, OntoLearn~\cite{ontolearn,ontolearn-reloaded} added syntactic parsing and graph-based pruning, and OntoGen~\cite{fortuna2007ontogen} introduced interactive user-in-the-loop refinement. These were principled systems, but often brittle. They struggled under terminological variation, struggled to align with established vocabularies, and produced ontologies too inconsistent to reuse reliably across domains. The second wave scaled up: NELL~\cite{carlson2010toward} continuously populated seed ontologies via semi-supervised learning, Klink-2~\cite{osborne2015klink} mined co-occurrences and citations to construct a 14,000-topic knowledge structure, and industry deployments, such as Microsoft Probase~\cite{wu2012probase}, Diffbot~\cite{diffbot}, and PoolParty~\cite{poolparty}, leveraged web-scale extraction and graph-based methods. However, increased scale did not compensate for the absence of logical rigor, often resulting in large but weakly structured knowledge collections. In parallel, the community built ontology support infrastructure, including BioPortal~\cite{noy2009bioportal,whetzel2011bioportal,bioportal_fairsharing_2024}, Linked Open Vocabularies (LOV)~\cite{vandenbussche2016linked}, Ontology Lookup Service (OLS)~\cite{jupp2015new,gatto2025rols}, the Open Biomedical Ontologies (OBO) Foundry~\cite{smith2007obo}, the TIB Terminology Service~\cite{Kraft:968600,tib_terminology_service,denbi_ontologies_terminology_2025}, Unified Medical Language System (UMLS)~\cite{bodenreider2004unified}, and FAIRsharing~\cite{sansone2019fairsharing}—essential repositories for hosting, discovering, reusing, and aligning ontologies across domains. Yet none was designed to evaluate or automate the learning process itself~\cite{kamdar2017systematic}. Then, with the emergence of large language models (LLMs), a new wave of ontology learning approaches has emerged: NeOn-GPT~\cite{fathallah2024neon} used structured prompts to elicit concept hierarchies, OntoGPT~\cite{caufield2024structured} applied schema-based extraction via recursive zero-shot prompting, OntoChat~\cite{zhang2024ontochat} introduced conversational interfaces for early modeling tasks, and OLLM~\cite{lo2024end} generated entire taxonomies end-to-end from domain corpora. Hybrid pipelines that combine embeddings with few-shot prompting outperformed single-method baselines~\cite{giglou2024llms4ol,giglou2025llms4ol}, and human-LLM collaboration reduced extraction effort by over 50\%~\cite{kommineni2024human}. Yet logical inconsistencies persisted~\cite{fathallah2024neon}, outputs remained domain-narrow~\cite{caufield2024structured}, semantic drift proved difficult to control~\cite{kommineni2024human}. Our early benchmarking efforts in \textsc{LLMs4OL} initiative~\cite{babaei2023llms4ol,giglou2024llms4ol,giglou2025llms4ol} and efforts to benchmark LLMs systematically on OL tasks confirmed this directly: foundation models can assist with OL tasks (i.e, term typing, taxonomy induction, and relation extraction), but no shared framework has been established to evaluate, compare, or build upon these capabilities across domains. Taken together, these limitations point to a broader gap: \textit{the absence of a unified infrastructure capable of grounding, integrating, and systematically evaluating automated OL across methods and domains}.

\begin{table*}[!tbh]
    \small
    \caption{Comparative positioning of OntoLearner.}
    \label{tab:combined_compare}
    \centering
    \begin{subtable}[t]{\textwidth}
    \centering
    \caption{Feature comparison of OntoLearner against major ontology infrastructure categories: OD (Ontology Development and Repository Platforms), OLS (Ontology Access and Lookup Services), and VI (Vocabulary and Terminology Integration Services). $\sim$ indicates planned functionality.}
    \label{tab:compare}
    \small
    \begin{tabular}{l|c|c|c|c}
    \hline
    \textbf{Feature} & \textbf{OD} & \textbf{OLS} & \textbf{VI} & \textbf{OntoLearner} \\
    \hline
    Cross-Domain Coverage              & \xmark & \cmark & \cmark & \cmark \\
    Ontology Hosting and Versioning    & \cmark & \xmark & \xmark & \cmark \\
    Search and Browse Functionality    & \cmark & \cmark & \cmark & \cmark \\
    Programmatic Access (API)          & \cmark & \cmark & \xmark & \cmark \\
    Vocabulary Alignment and Mapping   & \cmark & \xmark & \cmark & $\sim$ \\
    LLM Integration                    & \xmark & \xmark & \xmark & \cmark \\
    Ontology Editing/Authoring Tools   & \cmark & \xmark & \xmark & \xmark \\
    Modular and Reusable Components    & \xmark & \xmark & \xmark & \cmark \\
    Support for Ontology Enrichment    & \cmark & \xmark & \xmark & \cmark \\
    FAIR Principles Alignment          & \cmark & \cmark & \cmark & \cmark \\
    \hline
    \end{tabular}
    \end{subtable}

    \begin{subtable}[t]{\textwidth}
    \centering
    \caption{Comparison of OntoLearner with recent AI/LLM-based ontology learning systems. ``Bench.'' indicates provision of standardized benchmarks or dataset splits.}
    \label{tab:llm_systems}
    \small
    \resizebox{\textwidth}{!}{%
    \begin{tabular}{l|p{5cm}|p{5cm}|l|c}
    \hline
    \textbf{System} & \textbf{Main Tasks} & \textbf{LLM Integration} & \textbf{Domain Scope} & \textbf{Bench.} \\
    \hline
    NeOn-GPT~\cite{fathallah2024neon} & Requirements, class hierarchy, OWL axioms & GPT-3.5 prompting & Specific (Wine ontology) & \xmark \\
    OntoGPT~\cite{caufield2024structured} & Entity/relation extraction & Schema-driven zero-shot prompting & Biomedical, recipes & \xmark \\
    OntoChat~\cite{zhang2024ontochat} & Requirement engineering (e.g., CQs) & Interactive prompting & General (e.g., Music) & \xmark \\
    OLLM~\cite{lo2024end} & Taxonomy induction & Fine-tuned LLM & Cross-domain (Wikipedia, ArXiv) & \xmark \\
    Ontolearn~\cite{demir2025ontolearn} & OWL class expression learning & LLM-based verbalization (support module) & Large KGs (symbolic focus) & \xmark \\
    \textbf{OntoLearner (ours)} & Term typing, taxonomy, relations & Modular prompting, hybrid workflows & 22 domains (180 ontologies) & \cmark \\
    \hline
    \end{tabular}%
    }
    \end{subtable}
\end{table*}

The \autoref{tab:compare} summarizes and compares the strengths of each infrastructure category, whereas the \autoref{tab:llm_systems} contrasts AI/LLM-based learning systems. The pattern reveals a deeper problem than any single method could solve. Automation tools and ontology infrastructure evolved in isolation—one side building learners, the other building repositories, neither building the bridge between them. LLMs added generative power but no formal grounding. And critically, across all three waves, there was never a unified framework for measuring progress: no shared benchmarks, no standardized splits, no cross-domain evaluation. Without standardized evaluation, progress across iterations remained difficult to accumulate.

We introduce \textbf{OntoLearner}, a modular, cross-domain OL framework that addresses this gap by unifying, for the first time, ontology access, AI/LLM-driven learning, and systematic benchmarking into a single coherent infrastructure. OntoLearner builds upon the ontology infrastructure ecosystem, particularly BioPortal, OLS, OBO Foundry, LOV, and FAIRsharing, and, as a first-of-its-kind, releases 180 ontologies across 22 domains as machine-readable, pipeline-ready, and LLM-compatible datasets hosted on HuggingFace, directly bridging the gap between ontology repositories and modern AI pipelines. Unlike prior ontology hubs that focus on browsing and discovery, OntoLearner enables standardized loading, versioning, and streaming across Python workflows, making it well-suited for seamless integration with pretrained models, tokenizers, and evaluation scripts, supporting reproducible and scalable OL. Rather than replacing human expertise, OntoLearner positions LLMs as assistive components grounded in curated vocabularies, combining the rigor of symbolic knowledge engineering with the generalization power of foundation models, and providing the infrastructure to measure how well that combination actually works. Concretely, OntoLearner delivers four contributions: (1) a unified interface for transforming ontologies from diverse repositories into machine-readable formats suitable for LLM applications; (2) a modular library supporting core OL tasks (including term typing, taxonomy discovery, and non-taxonomic relation extraction) with standardized train/dev/test splits and evaluation metrics enabling rigorous, reproducible comparison across classical, neural, and hybrid methods; (3) AI/LLM integration as assistive components aligned with curated vocabularies, supporting concept suggestion, term typing, hierarchy completion, and definition generation grounded in existing knowledge models; and (4) an extensible, domain-agnostic platform aligned with FAIR principles, sustained through HuggingFace\footnote{\url{https://huggingface.co/collections/SciKnowOrg/ontolearner-benchmarking-6823bcd051300c210b7ef68a}}, GitHub\footnote{\url{https://github.com/sciknoworg/OntoLearner/}}, and complete documentation\footnote{\url{https://ontolearner.readthedocs.io/}} to ensure long-term accessibility and community-driven development.

OntoLearner is a standalone OL library rather than a wrapper around HuggingFace, that introduces high-level abstraction layers bridging symbolic and neural methods, with a long-term vision of becoming a standard library for OL, serving semantic web and machine learning researchers and practitioners. Beyond providing infrastructure, OntoLearner enables systematic empirical study of OL across domains. Using this framework, we show evidence of how classical and modern embedding models, as well as LLM-based approaches, behave differently across OL tasks, where semantic drift and logical inconsistencies emerge, and how grounding in curated ontologies affects these outcomes. These findings transform OntoLearner from a tooling contribution into a basis for reproducible scientific evaluation of OL methods, advancing the field from exploratory prototyping toward trustworthy, AI-assisted ontology engineering.

\section{Related Work}
Ontology learning (OL) has progressed from manual efforts to automated pipelines supported by specialized infrastructure, classical systems, and, more recently, LLMs. We organize related work into three themes: (1) ontology infrastructure and repositories for reuse and interoperability, (2) automated systems for extracting ontologies from text, and (3) the emerging role of LLMs. This framing highlights OntoLearner’s distinct role as a benchmarking framework that unites the three areas by enabling systematic evaluation and comparison across infrastructure, classical systems, and LLM-based methods.

\subsection{Ontology Infrastructure and Repositories}

A mature ecosystem of platforms supports ontology creation, discovery, and reuse across scientific domains. BioPortal, developed by the National Center for Biomedical Ontology (NCBO), exemplifies the ontology development and repository platform (OD category), hosting over 1,000 biomedical ontologies with rich search, visualization, and mapping tools \cite{noy2009bioportal,whetzel2011bioportal,bioportal_fairsharing_2024}. The Open Biomedical Ontologies (OBO) Foundry takes a community-driven approach, coordinating over 200 interoperable ontologies built on shared principles of openness and orthogonality to minimize redundancy \cite{smith2007obo}. Domain-specific portals like AgroPortal extend this model to agriculture and plant sciences by reusing the BioPortal infrastructure \cite{jonquet2018agroportal}.

Complementing development platforms are access and lookup services (OL category) that provide centralized, API-driven discovery. The EMBL-EBI Ontology Lookup Service (OLS) offers a unified REST API for querying around 270 OWL and OBO ontologies \cite{jupp2015new,gatto2025rols}, while the TIB Terminology Service extends this model beyond life sciences to engineering, chemistry, and physics domains \cite{Kraft:968600,tib_terminology_service,denbi_ontologies_terminology_2025}. Vocabulary integration services (VI category) focus on term reuse and alignment: the Linked Open Vocabularies (LOV) project curates metadata for commonly used RDF vocabularies \cite{vandenbussche2016linked}, while the Unified Medical Language System (UMLS) integrates over 60 biomedical terminologies into a single Metathesaurus with more than 2 million terms \cite{bodenreider2004unified}. Standards registries like FAIRsharing provide interlinked catalogs of community standards, databases, and ontologies to enhance visibility and FAIR compliance \cite{sansone2019fairsharing}.

While these platforms excel at hosting, discovery, or integration, the landscape remains fragmented with overlapping content and inconsistent modeling across ontologies \cite{kamdar2017systematic}. More critically, none provides systematic support for \textit{evaluating} how well AI models can learn, reconstruct, or enrich ontologies. Importantly, none provides end-to-end engineering support or systematic benchmarking of AI-driven ontology enrichment. \autoref{tab:compare} summarizes the strengths of each infrastructure category and highlights OntoLearner's novel contribution as an OL support system for benchmarking and evaluation.

\subsection{Classical Ontology Learning Frameworks}

Classical OL systems established foundational approaches by combining natural language processing (NLP) with machine learning (ML) in modular pipelines, prior to the advent of large language models. Text2Onto \cite{cimiano2005text2onto} pioneered probabilistic ontology modeling by integrating linguistic pattern matching (e.g., Hearst patterns), statistical association measures (PMI, TF-IDF), and hierarchical clustering to extract terms and taxonomic relations. It introduced confidence scoring, supported incremental learning from evolving corpora, and featured a graphical user interface for interactive refinement. Velardi et al.'s OntoLearn \cite{ontolearn} employed lexical patterns and syntactic parsing to induce taxonomies, with the Reloaded version \cite{ontolearn-reloaded} using graph-based pruning algorithms to construct coherent tree hierarchies, successfully reconstructing parts of WordNet. OntoGen \cite{fortuna2007ontogen} demonstrated early human-in-the-loop design by combining unsupervised k-means clustering and TF-IDF weighting to suggest candidate concepts through a user-friendly editor. Its plug-in architecture and interactive visualizations foreshadow many design choices in modern LLM-based ontology assistants.

Larger-scale systems pushed the boundaries of automation and coverage. The Never-Ending Language Learner (NELL) \cite{carlson2010toward} employed coupled semi-supervised learning using Naive Bayes classifiers, logistic regression, and bootstrapped extraction rules to continuously populate a seed ontology from web text. Over years, NELL accumulated millions of beliefs under knowledge coupling constraints to reduce semantic drift. However, it did not support the invention of new classes or complex axioms. Klink-2 constructed the 14,000-topic Computer Science Ontology by analyzing keyword co-occurrence and citation networks in scholarly publications \cite{osborne2015klink}. OLAF proposed a fully automated pipeline using dependency parsing, frequency heuristics, and rule-based relation extraction to produce ``minimum viable ontologies'' with minimal human intervention \cite{schaeffer2023olaf}.

In industry, OL has been deployed at massive scale. Microsoft's Probase mined 1.68 billion web pages using Hearst pattern extraction and statistical co-occurrence analysis to construct a 2.7-million-concept probabilistic taxonomy capturing relationship uncertainty \cite{wu2012probase}. Diffbot's commercial Knowledge Graph integrates computer vision, NLP, and entity resolution to extract structured data from over 60 billion web pages into a proprietary ontology with 10+ billion entities \cite{diffbot}. Commercial systems like PoolParty combine NLP-based term extraction, named entity recognition, and graph-based disambiguation in enterprise-grade taxonomy engineering platforms \cite{poolparty}.

These classical systems demonstrate the viability of automated ontology construction, but most relied on deterministic pattern matching, shallow statistical features, or domain-specific heuristics. Their modular architectures and algorithmic innovations remain relevant to OntoLearner in three ways: as \textit{baselines} for comparing LLM performance, as \textit{hybrid components} that can augment LLM outputs (e.g., graph pruning to improve structural coherence), and as sources of \textit{validation metrics} (e.g., graph-based quality measures). OntoLearner’s planned Text2Onto module\footnote{\url{https://ontolearner.readthedocs.io/package_reference/text2onto.html}} extends this lineage by generating synthetic training corpora from existing ontologies, enabling systematic evaluation of how well LLMs can replicate classical pattern-based extraction pipelines.

\subsection{LLMs for Ontology Learning}

The emergence of LLMs has introduced new paradigms for OL, offering flexible pipelines for concept extraction, taxonomy induction, and interactive modeling. However, recent evaluations reveal both promise and limitations. NeOn-GPT integrates structured prompt engineering sequences with the NeOn Methodology to elicit concepts, hierarchies, and OWL axioms. Evaluations on a gold-standard Wine ontology found that while the model could produce plausible structures, it struggled with logical consistency and reasoning, necessitating iterative human-in-the-loop refinement \cite{fathallah2024neon}. OntoGPT applies schema-based extraction via the SPIRES method in biomedical domains, using recursive zero-shot prompting grounded in existing ontology identifiers \cite{caufield2024structured}. OntoChat complements these by introducing conversational interfaces that parse intent and dialogue states to guide early-stage modeling tasks such as competency question generation \cite{zhang2024ontochat}.

A more radical approach is OLLM, which forgoes modular design in favor of end-to-end fine-tuning: a language model is trained to generate entire taxonomies from domain corpora using custom regularizers and embedding-based graph metrics. Compared to pipelined methods, OLLM achieves higher semantic coherence on Wikipedia and arXiv-derived ontologies \cite{lo2024end}. Complementary to these generative systems, Ontolearn~\cite{demir2025ontolearn} focuses on learning OWL class expressions over large-scale knowledge graphs. It implements symbolic and neuro-symbolic algorithms (e.g., EvoLearner \cite{heindorf2022evolearner}, DRILL \cite{demir2023drill}) to induce class expressions, and adds an LLM-powered verbalization module to translate those expressions into natural language. While not an end-to-end OL system, Ontolearn contributes important tooling for downstream reasoning, explanation, and integration with SPARQL-based infrastructure.

To address the lack of standardized evaluation settings across these systems, the LLMs4OL initiative we introduced formalizes the OL task space by benchmarking core subproblems—term typing, taxonomy discovery, and non-taxonomic relation extraction—across domains including medicine, geography, and lexicosemantics \cite{babaei2023llms4ol}. Challenge submissions showed that hybrid pipelines combining semantic embeddings (e.g., BERT, mpnet) with few-shot LLM prompting consistently outperform single-method baselines, though performance varies by domain and task \cite{giglou2024llms4ol,giglou2025llms4ol}. Studies of human–LLM collaboration report that LLMs can automate repetitive extraction steps and reduce effort by over 50\%, but semantic drift and inconsistencies persist without human oversight \cite{kommineni2024human}.

\autoref{tab:llm_systems} contrasts these LLM-based systems and highlights a key distinction: most are \textit{construction tools} focused on building ontologies, whereas OntoLearner is a \textit{benchmarking framework} designed to evaluate and compare OL methods. It directly implements and extends the LLMs4OL paradigm by providing standardized datasets, evaluation metrics, and reproducible pipelines across 177 ontologies in 22 domains—far exceeding prior efforts. This enables researchers to systematically assess methods such as NeOn-GPT’s structured prompting, OntoGPT’s schema-based extraction, retrieval-augmented generation, or OLLM’s fine-tuning.

OntoLearner distinguishes itself through three key contributions. First, its benchmarking-first design enables curated ontologies to be transformed into train/dev/test splits with automatic evaluation metrics, supporting rigorous model comparisons over ad hoc experimentation. Second, it bridges symbolic and neural approaches: classical algorithms serve as baselines; LLMs augment or replace modules; and hybrid workflows combine structured retrieval with generative modeling. Third, its modular, open-source architecture supports community-driven development under FAIR principles, with seamless integration of new ontologies via GitHub or HuggingFace. By advancing OL from exploratory prototyping to reproducible evaluation, OntoLearner provides essential infrastructure for trustworthy, AI-assisted ontology engineering.

\section{Methodology}
The development of OntoLearner was guided by a systematic requirements analysis of existing ontology engineering tools (see~\autoref{tab:compare}), which revealed key gaps, including limited cross-domain support, minimal LLM integration, and weak modularity and reusability. Insights from our prior empirical study on LLMs for OL~\cite{babaei2023llms4ol} and lessons learned from the LLMs4OL challenge series organization~\cite{giglou2024llms4ol,giglou2025llms4ol} shaped the framework’s requirements. This section outlines the requirement specification, then describes the software implementation and architectural compliance, and finally details the experimental setup for LLM analysis in OL.

\subsection{Requirements Specification}

Requirements guide the systematic development of a robust and high-quality library. The MASTER template by Chris Rupp and The SOPHISTs~\cite{Rupp2021,TheSophists2016} is used to specify both functional and non-functional requirements in a clear, consistent, complete, and testable manner. It employs the keywords “shall,” “should,” and “will” to indicate obligation levels: “shall” for mandatory requirements, “should” for recommended but optional ones, and “will” for planned future enhancements.

The documented requirements are listed in \autoref{tab:requirements}. Functional requirements (labeled ``F'') define the core capabilities of the OntoLearner Python library, comprising \textbf{14} items. These cover ontology management (F1–F4) for hosting and accessing cross-domain ontologies, ontology exploration (F5–F7), core OL tasks (F8–F11), learning models for LLM integration (F12–F13), and user customizations and interactions (F14). Moreover, non-functional requirements (denoted by ``N'') focus on quality aspects such as performance, usability, compatibility, and maintainability, which include \textbf{8} requirements, each defining the qualities of the framework, including performance (N1, N2, N3), usability (N4, N5), scalability (N6, N7), and maintainability (N8). Overall, the requirements are designed to encompass multiple objectives, as highlighted in \autoref{tab:compare}.

\begin{table*}[t]
    \centering
    \caption{Functional (F) and Non-Functional (N) requirements for the OntoLearner Python library, specified using the MASTER requirements template by Rupp and The SOPHISTs. The \textit{Proven Feature} (PF) column lists implemented functionalities; the \textit{Current Deficit} (CD) column indicates areas for future development.} 
    \label{tab:requirements}
    \resizebox{\textwidth}{!}{%
    \begin{tabular}{c| p{17cm} |c| c}
        \hline
         & \textbf{Requirement} & \textbf{PF} & \textbf{CD} \\ \hline
        \hline
        F1 & The library shall be able to load ontologies from various local file formats (e.g., OWL, XML, RDF, TTL). &  \cmark &  \\ \hline
        F2 & The library shall provide a Python import module to load ontologies and datasets automatically. &  \cmark &  \\ \hline
        F3 & The library shall provide access to an external repository on Huggingface for loading ontologies and task datasets. &  \cmark &  \\ \hline
        F4 & The library should adapt cross-domain ontologies. &  \cmark &  \\ \hline
        
        F5 & The library should allow users to measure and display ontological metrics for loaded ontologies. &  \cmark &  \\ \hline
        F6 & The library should allow users to create, view, and modify ontology metadata (e.g., name, domain, version, license). &  \cmark &  \\ \hline
        F7 & The library should allow users to browse through classes and relations of given ontologies. &  \cmark &  \\ \hline
        
        F8 & The library shall support the term-typing task for OL. &  \cmark &  \\ \hline
        F9 & The library shall support OL's taxonomy discovery task.  &  \cmark &  \\ \hline
        F10 & The library shall support the non-taxonomic relation extraction task of OL. &  \cmark &  \\ \hline
        F11 & The library will enable the Text2Onto (terminology and type extraction) task in OL.&  \cmark &  \\ \hline
        
        F12 & The library shall integrate with external learners (e.g., LLMs, retrieval models).   &  \cmark &  \\ \hline
        F13 & The library should enable users to customize existing learning models provided by the library, including their configuration parameters. &  \cmark  &  \\ \hline
        F14 & The library should allow users to customize learning tasks, including task definitions and configurations. & & \cmark \\ \hline
        F15 & The library should support multi-agent or ensemble learning strategies for OL tasks. & & \cmark \\ \hline
        F16 & The library should provide a graphical user interface for learners to interactively explore and execute ontologies. & & \cmark \\ \hline
        \hline 
        
        N1 & The library shall efficiently handle increasing amounts of data and concurrent users. & \cmark & \\ \hline
        N2 & The library should process large ontologies, including those with thousands of classes and properties, within a reasonable time frame. & \cmark & \\ \hline
        N3 & The library should optimize memory usage to efficiently manage large ontologies. & \cmark & \\ \hline
        
        N4 & The library shall be intuitive, easy to use, and extendible, with comprehensive user guides, examples, tutorials, and clear documentation.& \cmark  & \\ \hline
        N5 & The library will handle exceptions gracefully, providing the user with meaningful error messages. & \cmark &   \\ \hline
        
        N6 & The library should be compatible with major operating systems (Windows, macOS, Linux) and support multiple Python versions.  & \cmark & \\ \hline
        N7 &The library shall be provided as a Python package and be easily installable via PyPI.  & \cmark & \\  \hline
        
        N8 & The library should adhere to best practices for software development, including version control, modular design, code-level documentation, and automated testing. & \cmark & \\ \hline
        N9 & The library should isolate optional extensions (e.g., UI, advanced learners) from the core library to prevent coupling and regression. &  & \cmark \\ \hline
    \end{tabular}
    }
\end{table*}

The \textit{Proven Feature} (PF) column in \autoref{tab:requirements} lists implemented and verified features, while the \textit{Current Deficit} (CD) column highlights areas that need further development. This clear separation between completed features and identified gaps facilitates effective tracking of development progress and helps prioritize future improvements.

The OntoLearner library satisfies 13 of 16 functional and 8 of 9 non-functional requirements, reflecting a mature and practically usable system while leaving scope for further development. This high adherence demonstrates the tool’s maturity and its suitability for practical applications in OL and knowledge representation, while also indicating areas for further refinement to ensure continuous improvement. Nevertheless, the requirements were defined by the main developers based on the library’s design objectives and architecture decisions; thus, no formal agreement measures were applied. This approach is consistent with research-oriented software artifacts, in which requirements serve as a structured specification rather than the outcome of multi-stakeholder negotiation. To mitigate over-specialization and support future development, several requirements are intentionally abstract (e.g., F12–F13, N8) to support extensibility and future integration of additional AI components and learning interaction paradigms.


\begin{figure*}
    \centering
    \includegraphics[width=\linewidth]{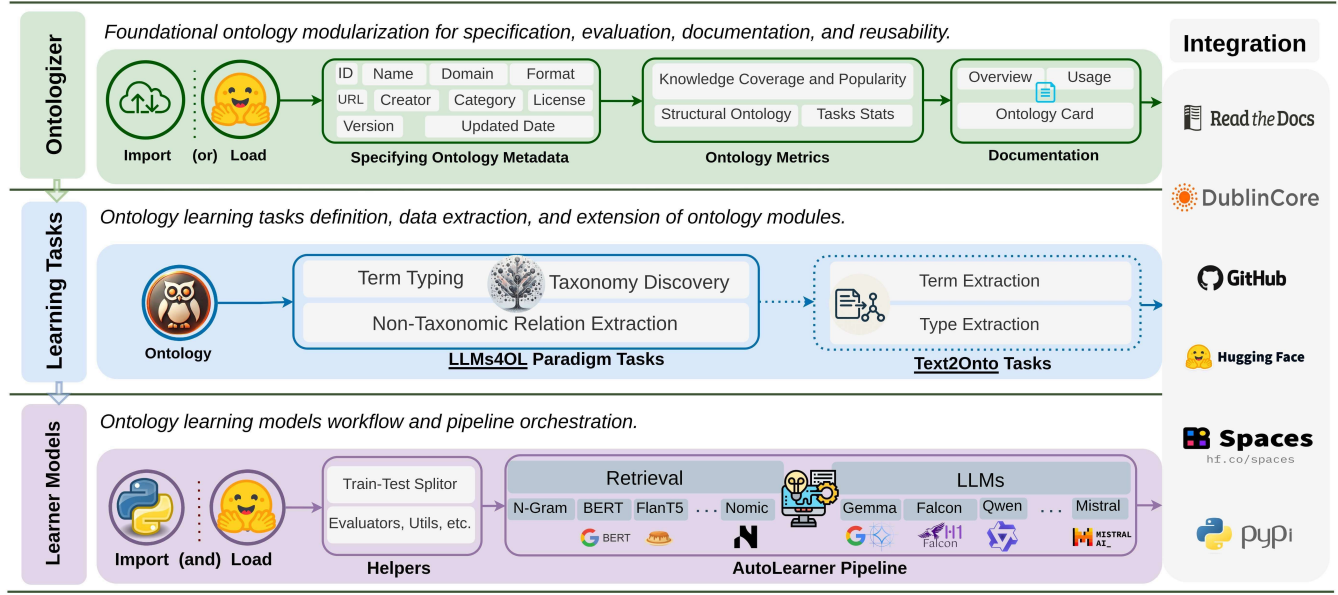}
    \caption{The conceptual and functional architecture of the OntoLearner library, illustrating its modular design for ontology access and learning within LLM workflows.}
    \label{fig:ontolearner-architecture}
\end{figure*}


\subsection{Software Core Components}
\label{sec:software-core-components}
OntoLearner is a modular Python framework for OL and reuse. It comprises three core components—Ontologizers, Learning Tasks, and Learner Models (see \autoref{fig:ontolearner-architecture}). In the following, we describe each core component in detail.

\sethlcolor{green!20}
\noindent\textbf{1. Ontologizer} -- \hl{\textit{A foundational module for ontology modularization that enables specification, evaluation, documentation, reuse, and seamless Pythonic import.}} As noted in the OD analysis in Related Work, domain-specific ontology repositories like BioPortal and AgroPortal promote alignment with community standards and research practices. To support similar integration across domains, the Ontologizer module offers an extensible Python interface for loading, inspecting, and repurposing ontologies within OntoLearner workflows. It transforms ontologies into programmatically accessible assets, enabling communities to adapt and reuse cross-domain ontologies (F4) as part of their workflows. This module offers the following key functionalities:
\begin{itemize}
    \item \textit{Processing}: To begin, inherit from the base ontology class~\footnote{\url{https://github.com/sciknoworg/OntoLearner/blob/main/ontolearner/base/ontology.py}}, which parses supported ontology file formats (OWL, RDF, XML, TTL) into a Pythonic object for programmatic manipulation. This ensures compatibility with Ontologizer’s core functions. If needed, predefined methods can be adjusted for specific cases—for example, ignoring blank nodes in certain learning tasks\footnote{Example fix: \url{https://github.com/sciknoworg/OntoLearner/blob/main/ontolearner/ontology/material\_science\_engineering.py\#L48}}.
    \item \textit{Import (or) Load}: Ontologies can be directly imported from the web or through Hugging Face Datasets (F3), streamlining the process of obtaining ontology data without manual file handling. It also supports loading data from the local machine (F1).
    \item \textit{Metadata Specification}:  Essential metadata attributes are defined for each ontology (i.e, ID, name, domain, format, version, creator, URL, license, category, update-date) (F6) to ensure traceability, version control, and compliance with FAIR principles. 
    \item \textit{Ontology Metrics Evaluation}: The module supports automated measurement of key ontology metrics (F5), such as knowledge coverage (i.e, number of classes, properties, individuals), structural complexity (i.e., hierarchical, breadth metrics), and popularity (i.e, nodes, edges, and leaf nodes). These metrics provide insights into the ontology's quality and suitability for various tasks.
    \item \textit{Automated Documentation}: Ontology documentation is generated following a Read-the-Docs style, providing clear, user-friendly insights into ontology usage, coverage, and structure (N4).
    \item \textit{Versioning}: Versioning supports the controlled evolution of ontology modules. Each ontology is assigned a version stored in its metadata, allowing users to track changes over time. Updates can occur in two ways: by submitting newer versions (e.g., via Hugging Face) or by proposing modifications through version control systems like GitHub (e.g., pull requests). These platforms enable collaborative development by tracking changes and facilitating community-driven improvements, ensuring alignment with evolving needs.
\end{itemize}
Once an ontology has been processed and integrated into OntoLearner, it can be imported using Python’s standard import mechanism (F2). This approach removes the need for manual downloads or complex loading procedures from external sources and provides users with easy-to-use functionality (N4). Moreover, as OntoLearner evolves, the ontology module will continuously be enhanced with new features to accommodate emerging requirements in ontology engineering, ensuring ongoing compatibility and functionality (N4, N8).

To support concurrent user involvement, this component maintains strict isolation between users and ontologies. The OL enables multiple simultaneous usage and updates, while subsequent modules rely solely on modularized ontologies for extension or learning strategies (N1). This design adheres to software engineering best practices (N8) and promotes scalable cross-domain ontology enrichment within OntoLearner. For example, we utilize multi-processing capabilities (N3) to optimize memory usage and efficiently manage large ontologies, even those with thousands of classes and properties (N2).

\sethlcolor{lightblue}
\noindent\textbf{2. Learning Tasks} -- \hl{\textit{The second step after Ontologizer is defining OL tasks for data curation.}} Within the OntoLearner framework, the modularized ontologies are extended with OL capabilities that automate core learning tasks. This extension is guided under the three core OL tasks we introduced via the LLMs4OL paradigm in 2023~\cite{babaei2023llms4ol}, which leverage LLMs to automate the key OL process (F8, F9, F10). The LLMs4OL paradigm is structured around three primary tasks essential for developing a primitive ontology: 1) \texttt{Term Typing}, the discovery of generalized type for a lexical term, 2) \texttt{Taxonomy Discovery}, the identification of the taxonomic hierarchy (is-a or subclass relationships) between type pairs, and 3) \texttt{Non-Taxonomic Relationship Extraction}, the extraction of non-taxonomic, semantic relations between types (relationships beyond ``is-a'').  By incorporating these tasks, OntoLearner ensures that its modularized ontologies, created with Ontologizer, are not only structured and reusable but also capable of continuous learning via automated OL techniques. 

Additionally, OntoLearner incorporates a preliminary implementation of the \texttt{Text2Onto} module \cite{cimiano2005text2onto}, which focuses on extracting ontological terms and types directly from raw text (F11). This component (described in more detail in \autoref{sec:text2onto}) is currently under active development and leverages LLMs for synthetic data generation to support term extraction and term typing, extending the LLMs4OL paradigm and enriching the framework’s learning objectives also to derive structured knowledge from unstructured sources. Notably, \texttt{Text2Onto} is designed to function independently of the LLMs4OL pipeline, ensuring modularity and preserving the integrity of existing components (N8). Users can import or load LLMs4OL tasks as inputs to \texttt{Text2Onto}, enabling flexible and extensible data extraction workflows for OL.

\sethlcolor{lightpurple}
\noindent\textbf{3. Learner Models} -- \hl{\textit{Enables workflow and pipeline orchestration of LLMs for OL, supporting seamless model execution and result exploration.}} The \textit{Learner Models} module serves as the orchestration layer of OntoLearner for managing OL workflows (F12, F13). It provides a flexible, extensible environment for designing, configuring, and experimenting with LLM-driven OL models. Key components include:
\begin{itemize}
    \item \textit{Machine Learning Utilities}: Core tools for robust modeling, such as train-test splits, label mappings, and evaluation metrics, are provided and easily importable (F12, N8). 
    \begin{itemize}
        \item \texttt{train\_test\_split}. The train-test split procedure is specifically designed to prevent data leakage across all OL tasks. Unlike a naive random split (e.g., directly applying Scikit-Learn \texttt{train\_test\_split}~\cite{scikitlearn}), OntoLearner implements a multi-stage, leakage-aware splitting algorithm including: 1) \textit{Term-Level Stratification} stage, where term types are first split using a stratified grouping strategy based on the least frequent associated type of each term. This balances rare and frequent types across splits. 2) \textit{Taxonomy Leakage Prevention} stage, in which taxonomic relations are then partitioned while enforcing that no term appearing in the training set is introduced into the test set. For relations not constrained by prior term assignment, the implementation leverages Scikit-Learn \texttt{train\_test\_split} to ensure statistically robust random sampling with reproducibility guarantees. 3) \textit{Relation-Type Preservation} stage, where non-taxonomic relations are split proportionally within each relation category to preserve semantic distribution across splits. 
        \item \texttt{LabelMapper}. A lightweight post-processing component that maps free-form LLM outputs to a fixed set of valid task labels. It was introduced because LLMs often generate lexical variations or loosely formatted answers (e.g., “true”, “Yes.”, “It is correct”), which can break evaluation pipelines. By normalizing such outputs into canonical labels, \texttt{LabelMapper} ensures consistent, valid, and evaluation-ready predictions in OL tasks.        
        \item \texttt{evaluation\_report}. A task-aware evaluation component that computes precision, recall, and F1-score for OL tasks using normalized pair- and triple-level matching. It ensures fair, consistent, and reproducible performance assessment, including proper handling of symmetric relations and lexical variations when required.
    \end{itemize}
    
     \item \textit{\texttt{Learner} Pipeline}:  This component orchestrates learner models into three primary categories: \texttt{AutoRetriever}, \texttt{AutoLLM}, and \texttt{AutoLearner}, following a modular and extensible design pattern (N8).

     \begin{itemize}
         \item \texttt{AutoRetriever}. This module plays a standard interface role for retrieval systems and supports a range of retrieval approaches, from traditional (e.g., NGrams) to advanced neural embeddings (e.g., BERT~\cite{devlin2019bert}, Nomic-embed~\cite{nussbaum2024nomic}), and enables search and navigation across ontology classes (F7). 

         \item \texttt{AutoLLM}. This module abstracts over a variety of LLMs—from compact models like Qwen2.5-0.5B~\cite{qwen2.5} to larger ones such as LLaMA~\cite{touvron2023llama} and DeepSeek~\cite{liu2024deepseek}—for ontology-related tasks. It provides a uniform interface for LLM-based operations on ontology tasks. 
        
         \item \texttt{AutoLearner}. This module acts as the core abstraction layer, defining a common interface for adapting diverse learner models. The \texttt{AutoRetriever} and \texttt{AutoLLM} components are intentionally designed to keep retrieval and LLM logic separate from learner models, following the Interface Segregation Principle (ISP)~\cite{noback2018interface}. This separation allows any learner to use a retriever or LLM backend provided to it, regardless of the internal implementation. Such a design enables seamless integration and switching between different modeling techniques without requiring re-implementation of existing learner logic. It also supports the Loose Coupling principle~\cite{orton1990loosely}, ensuring that learners do not need to know how the LLM or retriever is instantiated or managed internally. In line with extensibility principles~\cite{nicolajsen2025extensibility}, \texttt{AutoLearner} defines learning strategies that can produce specialized modules such as \texttt{AutoRetrieverLearner} (when an \texttt{AutoRetriever} is provided) and \texttt{AutoLLMLearner} (when an \texttt{AutoLLM} is provided). Furthermore, \texttt{AutoLLM} and \texttt{AutoRetriever} can be combined in a retrieval-augmented generation (RAG)~\cite{lewis2020retrieval,gao2023retrieval} setup through the \texttt{AutoRAGLearner} module. Overall, \texttt{AutoLearner} supports composition by allowing other learner categories to be integrated as long as they follow the specified interface or signature (F12, F14, N4, N8).
     \end{itemize}
\end{itemize}
To enhance prompt management in LLM workflows, OntoLearner includes an additional submodule, \texttt{AutoPrompt}, that allows users to define custom prompts (F14). If no prompt is provided, a predefined default is used. For instance, a user may combine a sentence-transformer model~\cite{sentencebert2019} from \texttt{AutoRetriever} with a smaller LLM like Qwen2.5-0.5B~\cite{qwen2.5} in a RAG configuration using a custom prompt. OntoLearner thus provides a user-friendly yet powerful environment for diverse learning scenarios (F14, N8).

\subsection{Synthetic Corpus Generation}
\label{sec:text2onto}

\begin{algorithm}[t]
    \caption{Text2Onto Synthetic Corpus Generation}
    \label{alg:text2onto}
    \small
    \begin{algorithmic}[1]
    \Require Ontology $\mathcal{O}$, Document size parameter $k$, LLM configuration $\theta$
    \Ensure Synthetic corpus $\mathcal{C}$
    \State Extract terms $T$, types $Y$, taxonomic relations $R$ from $\mathcal{O}$
    \State Partition $(T, Y)$ into subsets $\{P_1, P_2, \ldots, P_n\}$ using CMST-inspired algorithm with capacity $k$
    \For{each partition $P_i$}
        \State Generate CNL statements $S_i$ from $P_i$
        \State Paraphrase $S_i$ using LLM with configuration $\theta$ to obtain document $d_i$
        \State Annotate $d_i$ with ground truth terms and types
    \EndFor
    \State Aggregate all $d_i$ into corpus $\mathcal{C}$
    \State Split $\mathcal{C}$ into train, development, and test sets
    \end{algorithmic}
\end{algorithm}

In addition to the OntoLearner core components, it supports the \texttt{Text2Onto} corpus synthesis module for automated corpus generation.  The \texttt{Text2Onto} module is a key extension of OntoLearner. While the broader goal of OL is to generate ontologies from unstructured text, the absence of gold-standard corpora poses a significant challenge for benchmarking such pipelines. To bridge this gap, \texttt{Text2Onto} focuses on synthetic corpus construction—generating unstructured text aligned with existing ontologies—to enable controlled evaluation of ontology generation workflows. The module leverages recent advances in LLM-based synthetic data generation (SDG)~\cite{josifoski-etal-2023-exploiting,Zaitoun_Sagi_Peleg_2024} to produce documents annotated with ground-truth terms and types. The roadmap includes integrating multiple SDG algorithms that can simulate different characteristics of real-world text, such as homogeneous passages focused on a single type or heterogeneous passages spanning multiple types. The current implementation supports one such algorithm, detailed in Algorithm~\ref{alg:text2onto}, based on the following workflow:
\begin{enumerate}
    \item \textit{Ontology Data Extraction.} Utilizing OntoLearner’s APIs, \texttt{Text2Onto} retrieves ontology terms (individuals), types, taxonomic relations, and non-taxonomic relations, serving as input for subsequent processes.
    \item \textit{Term and Type Partitioning.} To control document size and ensure semantic coherence, the module partitions ontology data using an algorithm inspired by the Capacitated Minimum Spanning Tree Problem (CMST)~\cite{gouveia2000hierarchy}. This method groups closely related terms and types into connected subgraphs, promoting homogeneity within each generated document.
    \item \textit{Synthetic Text Generation.} The module employs a two-step approach inspired by~\cite{Zaitoun_Sagi_Peleg_2024}: first, ontology axioms are verbalized using Controlled Natural Language (CNL)~\cite{kuhn2014survey} based on predefined templates aligned with specific axiom patterns. For example, the axiom \texttt{:Capricciosa rdf:type :NamedPizza} is rendered as ``\textit{Capricciosa is a type of Named Pizza}'', using a template for the \texttt{rdf:type} relation. In the second step, the resulting CNL statements are paraphrased by an LLM to produce more natural, fluent, and contextually coherent text. For instance, a set of axioms about pizzas may be transformed into: ``\textit{Capricciosa, Four Seasons, and Napoletana are distinct varieties of pizza, each offering a unique combination of toppings balancing savoury and briny. All three share toppings such as anchovies, mozzarella and olives}''. 
\end{enumerate}

Users can tune LLM hyperparameters (e.g., model, decoding, temperature, and token limit) to control text style and diversity. Outputs can be evaluated using criteria such as coherence, readability, informativeness, and conciseness~\cite{10.1145/3677389.3702565}, and are partitioned into train and test splits for downstream tasks.

\subsection{Ontology Complexity Scorer}
\label{sec:ontology-complexity-scorer}

\begin{table*}[t]
    \caption{Ontology complexity scoring framework and robustness analysis.}
    \begin{subtable}{\textwidth}
    \caption{Interpretation of the ontology complexity score $\mathcal{C}_{score}$.}
    \label{tab:comp-score-interpretation}
    \centering
    \resizebox{\textwidth}{!}{%
    \begin{tabular}{l p{14cm}}
        \hline
        \textbf{Value} & \textbf{Interpretation} \\
        \hline
        \hline
        $0 < \mathcal{C}_{score} < 0.20$ 
        & Low-complexity ontology. Modest graph size, shallow depth, limited relations, and simple modelling patterns with little semantic richness. \\
        \hline
        $0.20 \le \mathcal{C}_{score} < 0.40$ 
        & Moderate-low complexity. Contains meaningful class coverage, some hierarchy development, and moderate breadth or relational structure. \\
        \hline
        $0.40 \le \mathcal{C}_{score} < 0.60$ 
        & Medium complexity. Ontology exhibits developed hierarchy, moderate-to-large graph structure, and richer semantic or relational diversity. \\
        \hline
        $0.60 \le \mathcal{C}_{score} < 0.80$ 
        & High complexity. Ontology is large, multi-layered, with substantial branching, semantic features, and diverse relationships. Indicates significant modeling depth. \\
        \hline
        $\mathcal{C}_{score} \ge 0.80$ 
        & Very high complexity. Ontology is extremely large-scale or densely structured, with deep hierarchies, high fan-out, and rich semantic or lexical content. Typically challenging for reasoning, maintenance, or LLM-based processing. \\
        \hline
    \end{tabular}
    }
    \end{subtable}

    \begin{subtable}{\textwidth}
    \centering
    \caption{Robustness analysis of the ontology complexity score under different configurations. Rank stability is measured using Spearman’s $\rho$ with respect to the original configuration. The columns correspond to: \textit{Weight Perturbations} is $\pm$10\% and $\pm$20\% perturbations of metric family (graph, coverage, hierarchy, breadth, and dataset) weights; \textit{Sigmoid Variation}, the variation of parameters $a \in [0.3, 0.5]$ and $b \in [5.5, 6.5]$; \textit{Normalizations}, the Min--Max and Z-score are alternative normalization schemes; and \textit{Uniform Weighting} is an equal weighting across all metric families.}
    \label{tab:complexity-sensitivity}
    \begin{tabular}{l ccccc}
    \hline
     & \textbf{Weight Perturbations} & \textbf{Sigmoid Variation} & \textbf{Normalizations}  & \textbf{Uniform Weighting} \\
    \hline
    \textit{Variants}           & 24 & 20 & 2 & 1 \\
    \textit{$\rho_{\min}$ (min $\rho$)}  & 0.9990 & 1.000 & 0.8557 & 0.9968  \\
    \textit{$\bar{\rho}$ (mean $\rho$)} & 0.9996 & 1.000 & 0.8564 & 0.9968 \\
    \hline
    \end{tabular}
    \end{subtable}

\end{table*}
The OntoLearner ontology complexity scorer assesses the complexity of an ontology using a set of ontological metrics. Its workflow is described below.

\noindent\textbf{Complexity Scorer.} As shown in \autoref{fig:ontolearner-architecture}, per ontology, Ontologizer computes key ontology metrics, including graph metrics $\mathcal{G}$ (i.e., total nodes, edges, root nodes, and leaf nodes), knowledge coverage $\mathcal{C}$ (i.e., classes, individuals, properties), hierarchical structure $\mathcal{H}$ (i.e., maximum, minimum, and average depth), breadth $\mathcal{B}$ (i.e., maximum, minimum, and average breadth), and dataset characteristics $\mathcal{D}$ (i.e., number of term types, taxonomic/non-taxonomic relations, and average terms per type). To summarize these high-dimensional metrics, we introduced a complexity score: a single, reproducible value summarizing an ontology's structural complexity. Rather than a standalone metric, complexity is treated as an emergent property arising from the interaction of multiple internal characteristics. This unified, absolute score relies solely on an ontology's intrinsic properties, ensuring a self-contained measure that facilitates consistent benchmarking without external dependencies. Formally, we define this score by first representing the ontology metrics as a vector representation of $x = \{ x_j | j \in \mathcal{M}\}$, where $\mathcal{M}$ is the set of all metrics: $\mathcal{M} = \{ \mathcal{G} \cup \mathcal{C} \cup \mathcal{H} \cup \mathcal{B} \cup \mathcal{D} \}$. The calculation of this complexity score involves the following key steps:

\begin{enumerate}
    \item \noindent\textit{Metric-level Normalization.} To obtain an absolute, ontology-intrinsic measure, we first apply a monotone, sublinear transformation to each metric $\mathcal{M}$. The transformation is done by using the natural logarithm of $\tilde{x}_{j} = log(1+x_j)$. The $log (1 + x)$ transform counts that span in high order of magnitude while preserving ordering and sensitivity at small scales. This has proven to be a logical decision for normalization due to the fact that all the metrics are mostly count-based metrics. Moreover, unlike the min-max or percentile normalization, the $log(1+x)$ is local to the ontology and therefore preserves the requirement that the complexity score be self-explanatory and reproducible. 

    \item \noindent\textit{Weighted Aggregation.} Normalized metric values are combined with metric weights $w_j$ that encode the relative importance of different metric families (graph, coverage, hierarchy, breadth, and dataset). The raw aggregated score $\mathcal{C}_{w}$ is defined as following:  $\mathcal{C}_{w} (\mathcal{M}) = \frac{\sum_{j=1}^{m} w_j \, \tilde{x}_{j}}{ \sum_{j=1}^{m} w_j} \in [0, 1]$, where the $w_j$ is a per-metric weight that are defined as follows:

\[
w_j \;=\;
\begin{cases}
0.30, & \text{if } j \in \mathcal{G} \;\;(\text{Graph})\\[4pt]
0.25, & \text{if } j \in \mathcal{C} \;\;(\text{Knowledge Coverage})\\[4pt]
0.10, & \text{if } j \in \mathcal{H} \;\;(\text{Hierarchy Structure})\\[4pt]
0.20, & \text{if } j \in \mathcal{B} \;\;(\text{Breadth})\\[4pt]
0.15, & \text{if } j \in \mathcal{D} \;\;(\text{Dataset Characteristics})
\end{cases}
\]
    \item \noindent\textit{Sigmoid Squashing to $[0, 1]$.} The weighted aggregation score of $\mathcal{C}_{w}$ is unbounded, because $log (1 + x)$ grows without bound. To produce a final complexity score in the range of $[0, 1]$ that is smooth and interpretable, we applied a logistic (sigmoid) mapping with fixed parameter $a$ (steepness) and $b$ (midpoint):
$$\mathcal{C}_{score} = \sigma_{a,b} (C_w (\mathcal{M})) = \frac{1}{1 + e^{(- a (C_w (\mathcal{M}) - b))}} $$

\end{enumerate}
The typical parameter choices used in this work are $a=0.4$, $b=6.0$ (these are fixed, domain-informed constants and do not depend on the dataset). This pipeline of $x_j \rightarrow log (1 + x_j) \rightarrow \mathcal{C}_{w}(\mathcal{M}) \rightarrow \sigma_{a,b}$ was chosen because it (i) compresses heavy-tailed count metrics while preserving order, (ii) uses only ontology local transformations, so scores are reproducible and most importantly independent, and (iii) produces a smooth bounded complexity score $\mathcal{C}_{score} \in [0, 1]$ whose midpoint and sensitivity can be set by interpretable parameters $b$ and $a$. \autoref{tab:comp-score-interpretation} represents the $\mathcal{C}_{score}$ interpretation in five scales.

\noindent\textbf{Complexity Scorer Robustness Analysis.}  To evaluate the robustness of the proposed ontology complexity score, we perform four complementary sensitivity assessments (see \autoref{tab:complexity-sensitivity}) using 180 ontologies.  First, we test \textit{weight perturbations} by varying each metric-family weight (graph, coverage, hierarchy, breadth, dataset) by $\pm 10\%$ and $\pm 20\%$ while re-normalizing the remaining weights. Second, we analyze \textit{sigmoid variation} by changing the bounding parameters over $a \in [0.3, 0.5]$ and $b \in [5.5, 6.5]$.  Third, we compare \textit{normalization schemes} by replacing the default $\log(1+x)$ transformation with Min--Max and Z-score normalization. Fourth, we evaluate a \textit{uniform weighting} baseline where all metric families are assigned equal weight. The results reveal a consistent and interpretable pattern. 

The robustness analysis shows that weight perturbations have a negligible impact on ranking ($\rho_{\min}=0.9990$, $\bar{\rho}=0.9996$), indicating that the score is not dominated by any single metric family.  Sigmoid variation is effectively rank-invariant ($\rho \approx 1.000$ for all tested settings), confirming that this stage mostly rescales values without changing comparative ordering. The near-perfect rank stability indicates that the aggregation structure dominates over parameter tuning, suggesting that the metric captures intrinsic structural signals rather than parameter-specific artifacts. Uniform weighting also remains highly consistent with the original formulation ($\rho=0.9968$), suggesting that domain-informed weights refine but do not fundamentally determine rankings.  In contrast, normalization choice is the main source of sensitivity (Min--Max / Z-score: $\rho_{\min}=0.8557$, $\bar{\rho}=0.8564$), implying that cross-ontology rank positions are more affected by dataset-dependent scaling than by weighting or sigmoid parameterization. This behavior is expected because Min--Max and Z-score are dependent on collections of ontologies, whereas $log(1+x)$ is pointwise and less dependent on the ontology collection. Overall, the complexity score is strongly robust to parametric and weighting choices, with normalization strategy being the primary methodological factor requiring explicit justification.

\subsection{OntoLearner Software Access}
OntoLearner is integrated with open-source platforms such as GitHub, HuggingFace, and PyPI, and is accompanied by comprehensive documentation on Read the Docs. These integrations promote transparency, reproducibility, and reuse of OL workflows across communities and domains.

\noindent\textbf{GitHub and PyPI.} OntoLearner’s source code is hosted on GitHub under the MIT License at \url{https://github.com/sciknoworg/OntoLearner}, with detailed contribution guidelines available at \url{https://github.com/sciknoworg/OntoLearner/blob/main/CONTRIBUTING.md}. The project is also published on PyPI for convenient distribution. An automated CI/CD pipeline supports streamlined releases across major operating systems (Windows, macOS, Linux) and multiple Python versions (N6).

\noindent\textbf{Documentation.} OntoLearner features extensive, well-structured documentation hosted on Read the Docs: \url{https://ontolearner.readthedocs.io/}. Key sections include: \textit{Getting Started} – installation and quickstart guide; \textit{Ontologizer} – instructions for ontology modularization, hosting, and adding new ontologies; \textit{Learneing Tasks} - supported OL tasks and usage of benchmarked ontologies; \textit{Learning Models} – guidance for using LLMs in learning tasks; \textit{Benchmarking} – provides comprehensive documentation for \textbf{180} modularized ontologies across \textbf{22} domains are currently added to OntoLearner (F4); \textit{Package References} – detailed API reference. Documentation is automatically updated through a CI/CD pipeline, ensuring it stays current without interfering with benchmarking or library releases.

\noindent\textbf{HuggingFace.} OntoLearner’s end-to-end benchmarking is supported via HuggingFace, leveraged as a centralized hub for storing ontologies and learning task datasets across domains. Repositories are publicly accessible at \url{https://huggingface.co/SciKnowOrg} under the MIT License (F3, F4, N1). The ontologies and datasets are typically stored in the widely used formats such as OWL, RDF, XML, or TTL and are fully compatible with OntoLearner's processing workflows. Moreover, the OL tasks datasets are stored in JSON. Users can upload their own ontologies to HuggingFace either through their personal account or by contributing directly to the \href{https://github.com/sciknoworg}{SciKnowOrg organization} page. Users can contribute by following the OntoLearner guidelines via \url{https://ontolearner.readthedocs.io/ontologizer/new_ontologies.html}, or by submitting a pull request with ontology files and metadata to the SciKnowOrg GitHub repository for review and integration.


\subsection{Scalability and FAIR Alignment}

\noindent\textbf{Scalability.} OntoLearner scales efficiently across large, multi-domain ontology collections. It currently integrates 180 ontologies across 22 domains, with an average of 3,401 classes per ontology—totaling 612,223 classes. The full pipeline, including parsing, importing, and dataset extraction for LLMs4OL tasks, completes in approximately 25.6 seconds per ontology, or 4.5 seconds overall. The Ontologizer component is designed to handle ontologies of all sizes—from small ontologies such as the Conference ontology~\cite{Conference} to large-scale ontologies such as ChEBI~\cite{ChEBI} with tens of thousands of classes—ensuring consistent performance regardless of scale.

\noindent\textbf{Adherence to FAIR Principles.} Ontologies in OntoLearner are richly annotated (e.g., domain, version, license, and provenance) and discoverable via the documentation and Ontologizer for streamlined navigation, ensuring that they are both human- and machine-discoverable (\textit{Findable}). They are openly hosted on widely used platforms such as HuggingFace and GitHub, enabling easy retrieval and verification (\textit{Accessible}). OntoLearner supports multiple ontology serialization standards (i.e., OWL, RDF, XML, and TTL), adopts community vocabularies, and provides programmatic usage for integration into third-party workflows, thereby promoting semantic compatibility across tools and domains (\textit{Interoperable}). Finally, modular packaging, inheritance-based ontology definitions, and Pythonic \texttt{import} mechanisms ensure seamless reuse in downstream applications. Clear licensing and documentation further encourage community contributions, solidifying long-term sustainability of ontologies within the ecosystem (\textit{Reusable}).

\noindent\textbf{OntoLearner Dublin Core Metadata.} The \texttt{OntoLearnerMetadataExporter}\footnote{Metadata documentation: \url{https://ontolearner.readthedocs.io/ontologizer/metadata.html}} component in OntoLearner is designed to automatically generate Dublin Core (DCMI)~\cite{weibel2000dublin} metadata for each ontology in the benchmarking suite. This adds another layer of compliance with FAIR principles. Moreover, this module outputs DCMI, Ontologizers metadata info in RDF/XML formats containing key fields--title, description, creator, license, format, version, last-updated date, domain, category, and download URL--compatible with Protégé, RDFLib, or Apache Jena. Metadata is organized under a top-level collection resource describing the entire OntoLearner benchmark, and generation is automated via CI/CD workflows triggered on each new PyPI release.

\subsection{Modular and Extensible Architecture}
\label{sec:modular-design-software-principles}

\begin{figure*}[!tb]
    \centering
    \small
    \begin{subfigure}{\textwidth}
        \centering
        \includegraphics[width=\linewidth]{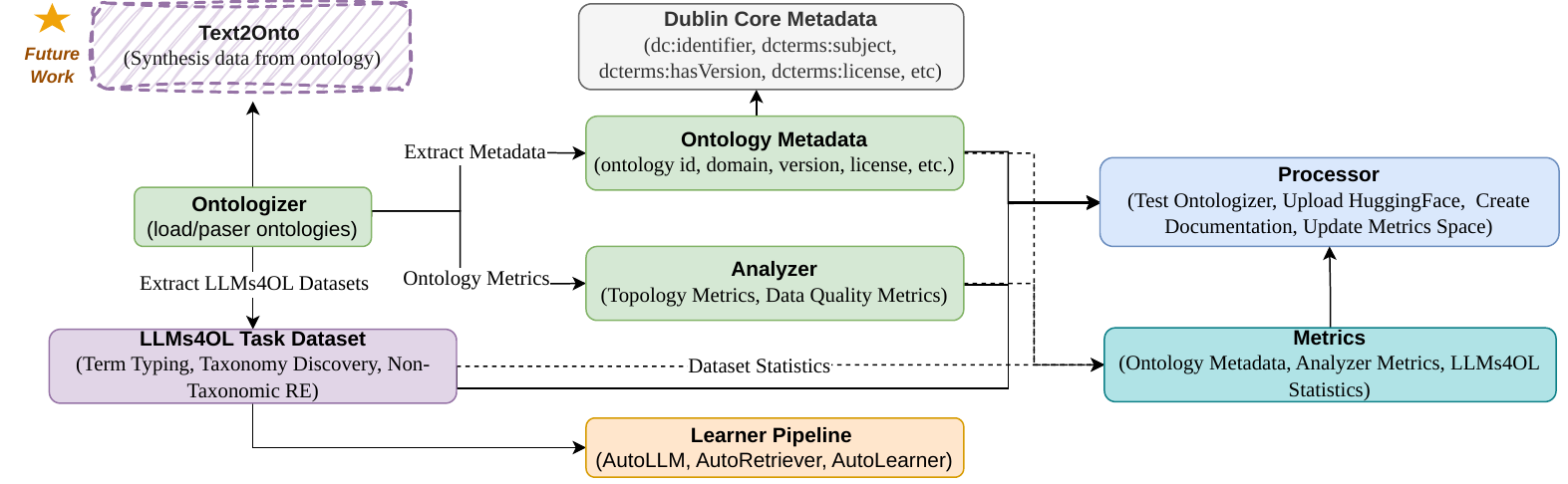}
        \caption{Architecture and module schema of the OntoLearner Python library. Planned future extensions are marked with a star.}
        \label{fig:schema-diagram}
    \end{subfigure}
    
    \begin{subfigure}{\textwidth}
        \centering
        \includegraphics[width=\linewidth]{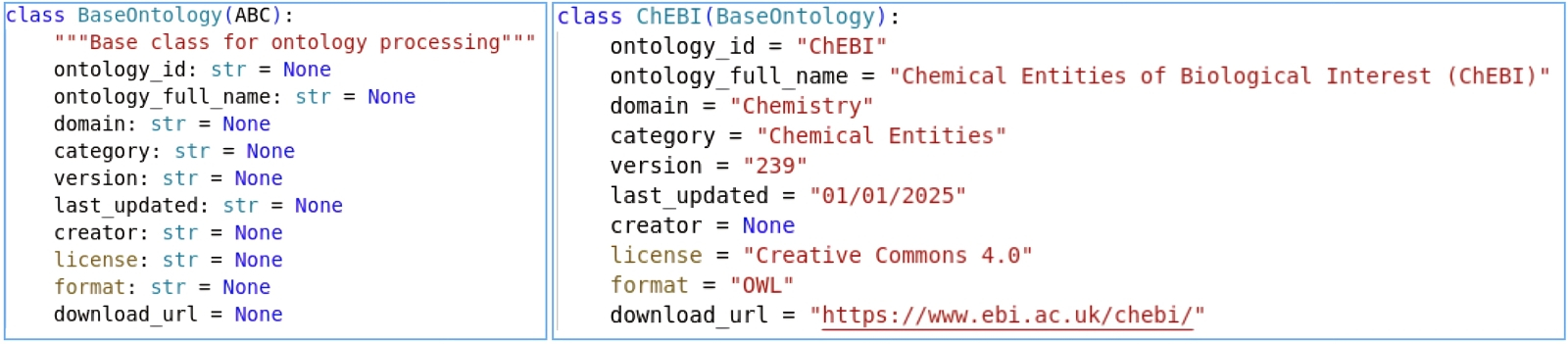}
        \caption{Illustrative example of ontology modularization within the Ontologizer component, showing a base ontology class alongside a modularized ChEBI ontology.}
        \label{fig:ontologizer-illustrative}
    \end{subfigure}
    
    \begin{subfigure}{\textwidth}
        \centering
        \includegraphics[width=\linewidth]{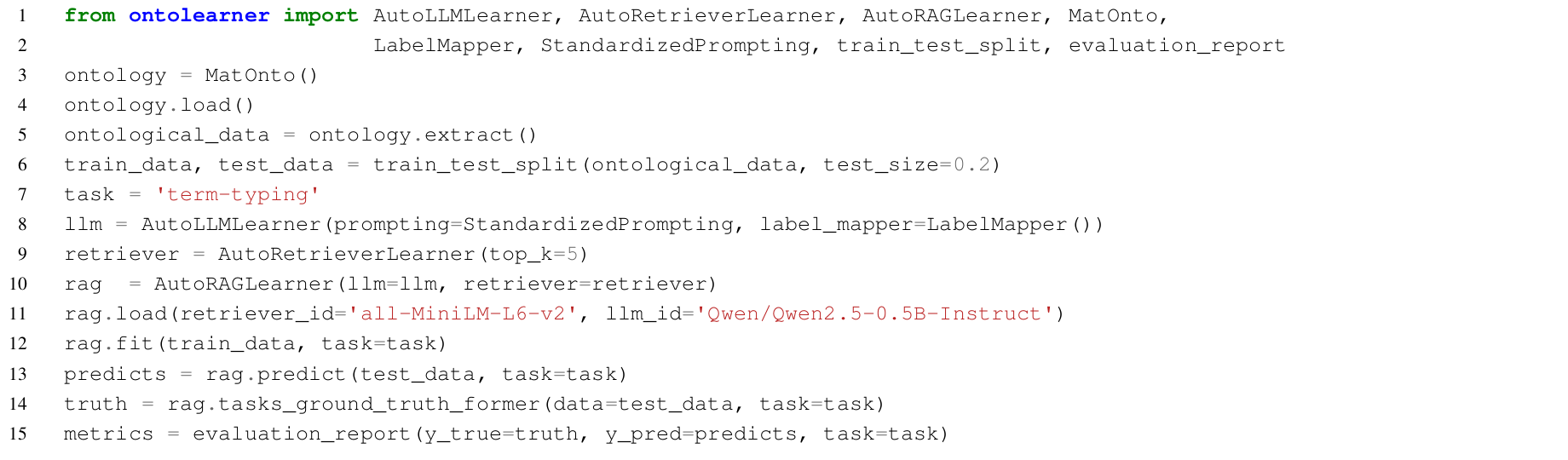}
    \caption{Step-by-step implementation of a RAG-based learner in OntoLearner for the \texttt{Term Typing} task from LLMs4OL. The same workflow applies to other LLMs4OL tasks by modifying the \texttt{task} parameter.}
    
    \label{lst:ontolearner-learner-code}
    \end{subfigure}

    \begin{subfigure}{\textwidth}
    \centering
    \includegraphics[width=\linewidth]{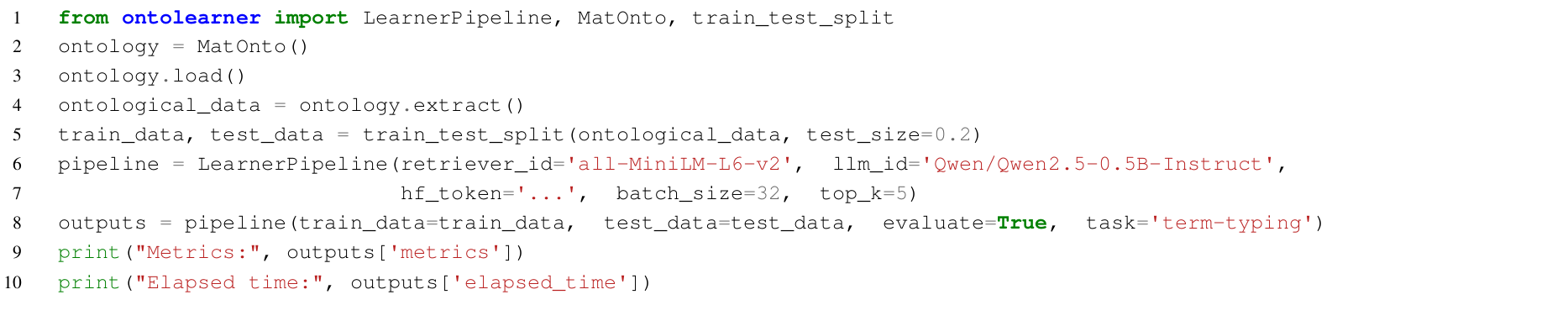}
    \caption{Example usage of the unified \texttt{LearnerPipeline} interface in OntoLearner for the \texttt{Term Typing} task, providing an end-to-end training and evaluation workflow.}
    
    \label{lst:ontolearner-learner-code-pipeline}
\end{subfigure}

    \caption{OntoLearner Architectural Design.}
    \label{fig:ontolearner-design-overview}
\end{figure*}

\noindent\textbf{Internal Component Interactions.} OntoLearner's high-level view is represented in \autoref{fig:ontolearner-architecture}, highlighting flexibility and scalability, enabling seamless integration of diverse ontologies through its modular and isolated architecture. This supports robust benchmarking and broad adaptability across ontological domains (F4) and end-to-end workflows. Moreover, the \autoref{fig:schema-diagram} shows internal components and their interactions. Ontologies are first ingested and parsed by the \texttt{Ontologizer}, which extracts structural and semantic metadata for further processing. The \texttt{Text2Onto} module 
can then generate synthetic training data from the parsed ontologies, supporting term/type extraction tasks of OL. This data, along with existing datasets, feeds into the \texttt{LearnerPipeline} for automated model configuration and learning. The \texttt{Analyzer} computes ontological metrics, while the \texttt{Processor} coordinates outputs: testing, documentation updates, and publishing to HuggingFace. Later, OntoLearner metadata will be exported with a collection of Ontologizers. Finally, all \texttt{Metrics} 
are consolidated to maintain an up-to-date benchmark space\footnote{Metric Space: \url{https://huggingface.co/spaces/SciKnowOrg/OntoLearner-Benchmark-Metrics}}, ensuring reproducibility and transparency across the library.

\noindent\textbf{Modular Design for Ontologizer Extensibility.} Ontologizer follows a modular design that simplifies ontology integration and supports extensibility with minimal overhead. As shown in \autoref{fig:ontologizer-illustrative}, a generic \texttt{BaseOntology} class defines a reusable metadata schema covering core attributes. Domain-specific ontologies, such as \texttt{ChEBI}, inherit from this base class and populate the required fields, enabling rapid and consistent onboarding of new ontologies\footnote{\url{https://github.com/sciknoworg/OntoLearner/blob/main/ontolearner/ontology/chemistry.py\#L39}}. This inheritance-based approach ensures that each ontology module remains self-contained, locally testable for metadata consistency, and easily integrable. Users can independently define, validate, and contribute modules via pull requests without altering core logic. This reduces duplication and complexity, enabling scalable, developer-friendly workflows aligned with sustainable software engineering principles. Moreover, the \texttt{train\_test\_split} used in \autoref{lst:ontolearner-learner-code} uses Ontologizer to apply a leakage-aware split partitioning strategy that ensures the evaluation results for all three OL tasks are not inflated by hidden structural overlaps between splits. This shows how Ontologizer's modular design is compatible with other components of OntoLearner.


\noindent\textbf{Software Principles Adaptations.} OntoLearner follows core software engineering principles (i.e., modularity, reusability, and separation of concerns) to provide a clear and extensible RAG framework. As illustrated in \autoref{lst:ontolearner-learner-code}, lines 3--5 handle ontology loading and data extraction, line 6 performs data splitting, line 8 configures the LLM and prompting, line 9 initializes the retriever, and lines 10--15 perform learner initialization, training, and evaluation. Each of these steps is implemented as an independent component adhering to the Single Responsibility Principle (SRP)~\cite{martin2006agile}. Retrievers (line 9) and LLMs (line 8) are pluggable through dependency injection, aligning with the Open/Closed Principle (OCP)~\cite{martin2006agile}, while learners depend only on the interfaces they require, following the Interface Segregation Principle (ISP)~\cite{hondros1977segregation}. Loose coupling~\cite{mammela2023loose} is achieved by keeping model and retrieval logic independent of learning strategies, enabling new techniques to be added without refactoring. As shown in \autoref{lst:ontolearner-learner-code-pipeline}, the \texttt{LearnerPipeline} (lines 6--8) further abstracts this workflow into a unified interface, reducing boilerplate code, promoting DRY (Don't Repeat Yourself) practices, and enabling reproducible experimentation across all LLM4OL tasks, similar to modern ML frameworks such as Transformers~\cite{wolf2020transformers}.

\noindent\textbf{Learner Adaptations.} OntoLearner, in addition to its core architecture, integrates learners  from teams participating in the LLM4OL challenge\footnote{Challenge Website: \url{https://sites.google.com/view/llms4ol}}(F12). In particular, learners from the top-performing teams \textit{RWTH-DBIS}~\cite{RWTH-DBIS}, \textit{SKH-NLP}~\cite{SKH-NLP}, \textit{Alexbek}~\cite{Alexbek}, and \textit{SBU-NLP}~\cite{SBU-NLP} have been incorporated to support benchmarking of diverse approaches. The \textit{RWTH-DBIS} learner applies knowledge-enhanced continual learning with prompt-tuned LLMs for term typing and taxonomy discovery. The \textit{SKH-NLP} learner combines BERT and LLaMA-based models for taxonomy discovery. The \textit{Alexbek} learner uses heterogeneous methods, including few-shot prompting, ensemble typing, and attention-based graph modeling, to address the full OL pipeline. The \textit{SBU-NLP} learner follows a training-free strategy using prompt engineering for stage-wise ontology construction. Further details on these learners are available in the OntoLearner documentation website\footnote{Learners at OntoLearner: \url{https://ontolearner.readthedocs.io/learners/llms4ol.html}}.

\section{Experimental Setups}

\subsection{Selected Benchmark Ontologies}

\begin{table*}[t]
    \caption{Benchmark ontologies for experimental analysis with their documentation page at \href{https://ontolearner.readthedocs.io}{ontolearner.readthedocs.io}.}
    \label{tab:benchmark-ontologies}
    \centering
    \resizebox{\textwidth}{!}{
    \begin{tabular}{l|l|l}
        \hline
        \textbf{ID} & \textbf{Ontology Full Name (Link)} & \textbf{Domain}\\
        \hline
        \hline
        \texttt{DoCO} & Document Components Ontology \cite{DoCO} (\href{https://ontolearner.readthedocs.io/benchmarking/education/doco.html}{doco}) & Education  \\
        \hline
        \texttt{Conference} & Conference Ontology \cite{Conference} (\href{https://ontolearner.readthedocs.io/benchmarking/events/conference.html}{conference}) & Events     \\
        \hline
         \texttt{GoodRelations} & Good Relations Language Reference \cite{GoodRelations} (\href{https://ontolearner.readthedocs.io/benchmarking/finance/goodrelations.html}{goodrelations})  & Finance \\
         \hline
         \texttt{Wine} & Wine Ontology \cite{Wine} (\href{https://ontolearner.readthedocs.io/benchmarking/food\_and\_beverage/wine.html}{wine})  & Food \& Beverage  \\
         \hline
         \texttt{AgrO} & Agronomy Ontology \cite{AgrO} (\href{https://ontolearner.readthedocs.io/benchmarking/agriculture/agro.html}{agro} ) &  Agriculture   \\
         \texttt{FoodOn} & Food Ontology \cite{FoodOn} (\href{https://ontolearner.readthedocs.io/benchmarking/agriculture/foodon.html}{foodon}) &   \\
         \hline
        \texttt{GeoNames} & GeoNames Ontology \cite{GeoNames} (\href{https://ontolearner.readthedocs.io/benchmarking/geography/geonames.html}{geonames}) & Geography  \\
         \hline
         \texttt{MaterialInformation} & Material Information Ontology \cite{MaterialInformation} (\href{https://ontolearner.readthedocs.io/benchmarking/materials\_science\_and\_engineering/materialinformation.html}{materialinformation}) & Materials Science \& Engineering   \\
        \texttt{MatOnto} & Material Ontology \cite{MatOnto} (\href{https://ontolearner.readthedocs.io/benchmarking/materials\_science\_and\_engineering/matonto.html}{matonto}) &    \\
        \texttt{PeriodicTable} & Periodic Table of the Elements Ontology \cite{PeriodicTable} (\href{https://ontolearner.readthedocs.io/benchmarking/materials\_science\_and\_engineering/periodictable.html}{periodictable}) &     \\
         \texttt{MDSOnto} & Materials Data Science Ontology \cite{MDSOnto} (\href{https://ontolearner.readthedocs.io/benchmarking/materials_science_and_engineering/mdsonto.html}{mdsonto}) &   \\
         \hline
         \texttt{GO} & Gene Ontology \cite{GO1,GO2} (\hyperlink{https://ontolearner.readthedocs.io/benchmarking/biology\_and\_life\_sciences/go.html}{go} ) & Biology \& Life Sciences \\
         \hline
         \texttt{ENVO} & Environment Ontology  \cite{ENVO1,ENVO2} (\href{https://ontolearner.readthedocs.io/benchmarking/ecology\_and\_environment/envo.html}{envo})& Ecology \& Environment  \\
          \texttt{SWEET} & Semantic Web for Earth and Environment Technology Ontology \cite{SWEET} (\href{https://ontolearner.readthedocs.io/benchmarking/ecology\_and\_environment/sweet.html}{sweet}) &    \\
         \hline
         \texttt{CCO} & Common Core Ontologies \cite{CCO} (\href{https://ontolearner.readthedocs.io/benchmarking/general\_knowledge/cco.html}{cco}) & General Knowledge\\
         \texttt{DBpedia} & DBpedia Ontology \cite{DBpedia} (\href{https://ontolearner.readthedocs.io/benchmarking/general\_knowledge/dbpedia.html}{dbpedia}) &  \\
         \texttt{SchemaOrg} & Schema.org Ontology \cite{SchemaOrg}  (\href{https://ontolearner.readthedocs.io/benchmarking/general\_knowledge/schemaorg.html}{schemaorg}) &    \\
         \hline
         \texttt{AUTO} & Automotive Ontology \cite{AUTO} (\href{https://ontolearner.readthedocs.io/benchmarking/industry/auto.html}{auto}) &  Industry \\
          \texttt{PTO} & Product Types Ontology \cite{PTO} (\href{https://ontolearner.readthedocs.io/benchmarking/industry/pto.html}{pto}) &   \\
         \hline
         \texttt{DOID} & Human Disease Ontology  \cite{DOID} (\href{https://ontolearner.readthedocs.io/benchmarking/medicine/doid.html}{doid}) & Medicine  \\
          \texttt{OBI} & Ontology for Biomedical Investigations \cite{OBI} (\href{https://ontolearner.readthedocs.io/benchmarking/medicine/obi.html}{obi}) &    \\
         \hline
         \texttt{OM} & Ontology of Units of Measure \cite{OM} (\href{https://ontolearner.readthedocs.io/benchmarking/units\_and\_measurements/om.html}{om}) & Units and Measurements   \\
          \texttt{QUDT} & Quantities, Units, Dimensions and Data Types \cite{QUDT} (\href{https://ontolearner.readthedocs.io/benchmarking/units\_and\_measurements/qudt.html}{qudt}) &    \\
         \hline
         \texttt{PROCO} & PROcess Chemistry Ontology \cite{PROCO1,PROCO2} (\href{https://ontolearner.readthedocs.io/benchmarking/chemistry/proco.html}{proco}) & Chemistry   \\
          \texttt{VIBSO} & Vibrational Spectroscopy Ontology \cite{VIBSO} (\href{https://ontolearner.readthedocs.io/benchmarking/chemistry/vibso.html}{vibso}) &    \\
          \texttt{ChEBI} & CHEBI Integrated Role Ontology  \cite{ChEBI} (\href{https://ontolearner.readthedocs.io/benchmarking/chemistry/chebi.html}{chebi}) &    \\
         \hline
    \end{tabular}
    }
    
\end{table*}

To ensure a comprehensive and unbiased evaluation, we selected a \textbf{26} diverse set of benchmark ontologies spanning multiple domains as summarized in \autoref{tab:benchmark-ontologies}. The selection was guided by three main criteria: domain diversity, structural and semantic heterogeneity, and community relevance. The benchmark set covers a broad spectrum of application domains, including education, events, finance, food and agriculture, geography, materials science and engineering, biology and life sciences, ecology and environment, medicine, chemistry, industry, and general knowledge. This diversity allows us to assess the robustness and generalizability of the evaluated methods across fundamentally different conceptual domains, ranging from highly specialized scientific ontologies (e.g., Gene Ontology~\cite{GO1}, ChEBI~\cite{ChEBI}, OBI~\cite{OBI}) to cross-domain and lightweight vocabularies (e.g., \href{https://schema.org/}{Schema.org}). Moreover, the selected ontologies exhibit substantial variation in size, modeling granularity, and logical complexity. Some ontologies, such as GeoNames~\cite{GeoNames} and PeriodicTable~\cite{PeriodicTable}, primarily provide taxonomic or descriptive structures with limited axiomatization, while others, such as GoodRelations~\cite{GoodRelations}, AgrO~\cite{AgrO}, FoodOn~\cite{FoodOn}, MDSOnto~\cite{MDSOnto}, ChEBI, and the Common Core~\cite{CCO} ontologies, make extensive use of formal constraints, rich class hierarchies, and object property semantics. This heterogeneity is essential for evaluating how well the approaches scale and adapt to different ontology engineering practices and levels of expressivity. 

For domains where multiple complementary ontologies exist (e.g., agriculture, materials science, ecology, medicine, chemistry, and units of measurement), we deliberately included more than one ontology to capture intra-domain variability. This enables a finer analysis of domain-specific effects and reduces the risk of concluding isolated modeling choices. Overall, the benchmark comprises 26 of the 180 ontologies from the OntoLearner collection, yielding a balanced and representative testbed for cross-domain experimental analysis.

\subsection{Selected Retrievers and Large Language Models}
\label{sec:selected-retrievers-and-llms}

\begin{itemize}[label={}, leftmargin=0pt]
    \item \noindent\textbf{Retrievers.} The $f(.)$ is an encoder for representing a query $Q$ and knowledge base (terms, types, and relationships) in a high-dimensional space. OntoLearner supports a diverse set of retrievers spanning lexical, embedding-based, and hybrid architecture retrieval models to enable fine-grained selections depending on the task at hand. 
    \begin{itemize}
        \item \textit{Lexical Retrievers.} We include several lexical baselines based on n-gram matching. These consist of word-level unigram frequency counting (Ngram-W-Counter) and word-level TF-IDF with sublinear term-frequency scaling ($1+\log(tf)$) (Ngram-W-TFIDF). In addition, we employ character-level 3-gram frequency counting (Ngram-C-Counter) and character-level 3-gram TF-IDF with sublinear scaling (Ngram-C-TFIDF). To capture mixed lexical patterns, we further include character–word 3-gram frequency counting (Ngram-CW-Counter) and character–word 3-gram TF-IDF (Ngram-CW-TFIDF).
        
        \item \textit{Traditional Word Embeddings. } We evaluate classical static embedding models, including GloVe~\cite{GloVe} using the \href{https://nlp.stanford.edu/projects/glove/}{Stanford GloVe} Dolma model trained on 220B tokens with a 1.2M vocabulary and 300-dimensional uncased vectors, as well as Word2Vec~\cite{Word2Vec} using the \href{https://code.google.com/archive/p/word2vec/}{Google Word2Vec} embeddings with 300-dimensional vectors.
        
        \item \textit{Sentence Transformers.} We include multiple sentence-level bi-encoder models from the Sentence Transformers framework~\cite{reimers-2019-sentence-bert}. MiniLM-L6 is a lightweight general-purpose bi-encoder trained on over one billion sentence pairs to produce normalized 384-dimensional embeddings (\href{https://huggingface.co/sentence-transformers/all-MiniLM-L6-v2}{all-MiniLM-L6-v2}). MPNET is a higher-capacity bi-encoder based on the MPNet architecture with mean pooling, trained on over one billion text pairs to generate normalized 768-dimensional embeddings (\href{https://huggingface.co/sentence-transformers/all-mpnet-base-v2}{all-mpnet-base-v2}). GTR-T5~\cite{GTR-T5} is a dual-encoder architecture trained specifically for semantic search, producing 768-dimensional dense representations (\href{https://huggingface.co/sentence-transformers/gtr-t5-base}{gtr-t5-base}). We also include domain-specific models, namely MatSciBERT~\cite{MatSciBERT}, a BERT model trained on materials science literature for improved information extraction (\href{https://huggingface.co/m3rg-iitd/matscibert}{matscibert}), and BiomedBERT~\cite{BiomedBERT}, a BERT-based model pretrained from scratch on PubMed abstracts that achieves state-of-the-art performance on biomedical NLP benchmarks (\href{https://huggingface.co/microsoft/BiomedNLP-BiomedBERT-base-uncased-abstract}{BiomedNLP-BiomedBERT-base-uncased-abstract}).
        
        \item \textit{Modern Embedding Models (Bi-Encoders).} We further evaluate modern large-scale embedding models. Nomic~\cite{Nomic} is an open-source text embedding model with long-context support that outperforms OpenAI Ada-002 on both short- and long-context benchmarks (\href{https://huggingface.co/nomic-ai/nomic-embed-text-v1.5}{nomic-embed-text-v1.5}). Nomic-MoE~\cite{Nomic-MoE} extends this approach with a Mixture-of-Experts architecture, improving efficiency while maintaining competitive performance (\href{https://huggingface.co/nomic-ai/nomic-embed-text-v2-moe}{nomic-embed-text-v2-moe}). We also include Qwen3 Embeddings~\cite{zhang2025qwen3}, a family of proprietary multilingual embedding models available in 0.6B, 4B, and 8B parameter sizes (\href{https://huggingface.co/Qwen/Qwen3-Embedding-0.6B}{Qwen3-0.6B}, \href{https://huggingface.co/Qwen/Qwen3-Embedding-4B}{Qwen3-4B}, and \href{https://huggingface.co/Qwen/Qwen3-Embedding-8B}{Qwen3-8B}). These models support long-context understanding, flexible embedding dimensions, and achieve state-of-the-art performance on benchmarks such as MTEB~\cite{muennighoff2022mteb}. Finally, we include Gemma~\cite{Gemma}, a lightweight multilingual embedding model optimized for on-device inference, supporting over 100 languages with low latency and efficient memory usage (\href{https://huggingface.co/google/embeddinggemma-300m}{embeddinggemma-300m}).
        
        \item \textit{Hybrid Cross-Encoder and Bi-Encoder Retrieval.} To balance efficiency and accuracy, we include a cascaded retrieval system, CS-Qwen3-8B-MiniLM~\cite{reimers-2019-sentence-bert}. This hybrid approach combines a \href{https://huggingface.co/Qwen/Qwen3-Embedding-8B}{Qwen3-Embedding-8B} bi-encoder for efficient candidate retrieval with a \href{https://huggingface.co/cross-encoder/ms-marco-MiniLM-L12-v2}{MiniLM-L12} cross-encoder fine-tuned on MS MARCO for accurate reranking, enabling scalable yet precise retrieval.

    \end{itemize}
    
    \item \noindent\textbf{Large Language Models. } We conducted experiments using a diverse collection of LLMs spanning multiple model families, parameter scales, and training paradigms, comprising a total of \textbf{12} models. From the Qwen family~\cite{yang2025qwen3}, we evaluated base, instruction-tuned, and reasoning-oriented variants, including \href{https://huggingface.co/Qwen/Qwen3-0.6B}{Qwen3-0.6B}, \href{https://huggingface.co/Qwen/Qwen3-4B-Instruct-2507}{Qwen3-4B-Instruct-2507}, \href{https://huggingface.co/Qwen/Qwen3-4B-Thinking-2507}{Qwen3-4B-Thinking-2507}, \href{https://huggingface.co/Qwen/Qwen3-8B}{Qwen3-8B}, \href{https://huggingface.co/Qwen/Qwen3-14B}{Qwen3-14B}, and the large-scale mixture-of-experts model \href{https://huggingface.co/Qwen/Qwen3-Next-80B-A3B-Instruct}{Qwen3-Next-80B-A3B-Instruct}. Similarly, from the Gemma family~\cite{kamath2025gemma}, we included instruction-tuned models at varying scales, namely \href{https://huggingface.co/google/gemma-3-1b-it}{Gemma-3-1b-it}, \href{https://huggingface.co/google/gemma-3-4b-it}{Gemma-3-4b-it}, \href{https://huggingface.co/google/gemma-3-12b-it}{Gemma-3-12b-it}, and \href{https://huggingface.co/google/gemma-3-27b-it}{Gemma-3-27b-it}.  In addition, we evaluated the instruction-tuned \href{https://huggingface.co/mistralai/Mistral-Small-3.2-24B-Instruct-2506}{Mistral-Small-3.2-24B-Instruct-2506} model~\cite{jiang2023mistral7b}, as well as a model from the Falcon family~\cite{zuo2025falcon}, specifically \href{https://huggingface.co/tiiuae/Falcon-H1-1.5B-Deep-Instruct}{Falcon-H1-1.5B-Deep-Instruct}. This comprehensive selection enables an in-depth analysis of the effects of model family, scale, and training strategy (e.g., instruction tuning and explicit reasoning objectives). 
\end{itemize}

\subsection{Retrieval-Augmented Generation Pipeline} 
The RAG module consists of two main components: a retriever and an LLM-based re-ranker. Let $Q$ denote an input query, the retriever defines a scoring function 
$s: Q \times \mathcal{C} \rightarrow \mathbb{R}$, where $\mathcal{C} = \{c_1, c_2, ..., c_N\}$ is the set of candidates. It selects the top-$k$ most relevant candidates as $\mathcal{C}_k(Q) = \operatorname{TopK}_{c \in \mathcal{C}} \; s(Q, c)$. Given $\mathcal{C}_k(Q)$, the LLM estimates the relevance of each candidate via $LLM_r : Q \times \mathcal{C} \rightarrow [0,1]$, and the final selection is $\mathcal{C}^*(Q) = \{ c \in \mathcal{C}_k(Q) \mid LLM_r(Q,c) = 1\}$. The scoring function uses cosine similarity, $s(Q, c) = \cos(q, c_d) = \frac{q^\top c_d}{\|q\|_2 \, \|c_d\|_2}$, where the query $Q$ and a candidate document $d \in \mathcal{D}$ are mapped to a shared embedding space using an encoder $q = f(Q)$ and $c_d = f(d)$. Retrieval performance, therefore, depends on the geometric alignment between query and candidate embeddings~\footnote{A single executable pipeline that instantiates all configurations described in this section is provided in the OntoLearner repository \url{https://github.com/sciknoworg/OntoLearner/blob/main/examples/pipeline.ipynb} and through the documentation website.}.

\noindent\textbf{Term Typing.} This task is defined as: \textit{Given a lexical term, identify its types}. Let a lexical term $\mathcal{L}$ be the query and $\mathcal{T}$ the set of candidate types. The retriever selects $\mathcal{T}_k(\mathcal{L}) = \operatorname{TopK}_{t \in \mathcal{T}} \; s(\mathcal{L}, t)$,
where $\mathcal{T} = \{t_1, t_2, ..., t_N \}$. Since a term may belong to multiple types, the LLM selects $\mathcal{T}^* (\mathcal{L}) = \{t \in \mathcal{T}_k(\mathcal{L}) \mid LLM_r (\mathcal{L}, t) = 1 \}$. 

\noindent\textbf{Type Taxonomy Discovery.} This task involves ``\textit{Given the term types, discover the taxonomy hierarchy between them}''. It focuses on constructing hierarchical relationships between types (i.e., is-a relationships). The OntoLearner supports two types of retrievers, described as follows:
\begin{itemize}
    \item \textit{Traditional Retriever.} Consider a taxonomy $\mathcal{T}$ represented as a directed acyclic graph $\mathcal{H} = (\mathcal{T}, \epsilon)$, where $(t_i, t_j) \in \epsilon$ indicates $t_i \sqsubseteq t_j$. In the RAG pipeline, taxonomy discovery is formulated as pairwise hierarchical relation prediction. Given $t_q \in \mathcal{T}$, the retriever scores all types as potential parent candidates using $s: \mathcal{T} \times \mathcal{T} \rightarrow \mathbb{R}$ and selects the top-$k$ candidates $\mathcal{T}_k(t_q) = \operatorname{TopK}_{t \in \mathcal{T}} s(t_q, t)$.  The LLM then acts as a hierarchical verifier, $LLM_h: \mathcal{T} \times \mathcal{T} \rightarrow [0,1]$, where $LLM_h(t_q, t_p) = 1$ indicates that $t_q$ is a subclass of $t_p$. The final predicted parent types are $\mathcal{P}^*(t_q) = \{ t_p \in \mathcal{T}_k(t_q) \mid LLM_h(t_q, t_p) = 1 \}$. Aggregating across all query types yields the discovered taxonomy $\epsilon^* = \bigcup_{t_q \in \mathcal{T}} \{(t_q, t_p) \mid t_p \in \mathcal{P}^*(t_q)\}, \quad \mathcal{H}^* = (\mathcal{T}, \epsilon^*)$. Following the OntoLearner architecture, this task reuses the term typing formulation, with $t_q \in \mathcal{T}$ as the query instead of $\mathcal{L}$. 
    
    \item \textit{LLM-Augmented Retriever.} This approach extends traditional retrievers by introducing an LLM-driven candidate generation stage prior to retrieval. Given $t_q \in \mathcal{T}$, the LLM first infers a coarse set of plausible parent candidates using generative AI. Formally, the LLM defines a candidate proposal function $LLM_c : \mathcal{T} \rightarrow \mathcal{T}_n$, producing a top-$n$ set $\mathcal{T}_n^{LLM}(t_q) = LLM_c(t_q)$, where $\lvert \mathcal{T}_n^{LLM}(t_q) \rvert = n$, and each $t_p \in \mathcal{T}_n^{LLM}(t_q)$ is a potential $(t_q, t_p)$ \textit{is-a} relation. The retriever then operates on $\mathcal{T}_n^{LLM}(t_q)$ rather than the full type set. Using the scoring function $s : \mathcal{T} \times \mathcal{T} \rightarrow \mathbb{R}$, the top-$k$ candidates are $\mathcal{T}_k(t_q) = \operatorname{TopK}_{t \in \mathcal{T}_n^{LLM}(t_q)} s(t_q, t)$, and the hierarchical verifier $LLM_h: \mathcal{T} \times \mathcal{T} \rightarrow [0,1]$ determines valid subclass relations using $\mathcal{P}^*(t_q) = \{ t_p \in \mathcal{T}_k(t_q) \mid LLM_h(t_q, t_p) = 1 \}$. By constraining the candidate space, this approach improves semantic alignment while preserving the modularity of the RAG-based taxonomy discovery pipeline. 
\end{itemize}

\noindent\textbf{Non-Taxonomic Relation Extraction.}  This task is defined as: \textit{Given two types, identify if there is a non-taxonomic semantic relationship between them, and if so, which one}. Unlike taxonomy discovery, which focuses on hierarchical \textit{is-a} relations, this task captures other meaningful associations (e.g., \textit{part-of}, \textit{used-for}, \textit{causes}, \textit{related-to}) to enrich the ontology. Let $\mathcal{R}$ denote the set of predefined non-taxonomic relation types. The task is formulated as a multi-stage RAG-based process. First, candidate type pairs are identified using the taxonomy discovery formulation: for $t_q \in \mathcal{T}$, the retriever selects $\mathcal{T}_k(t_q)$, yielding type pairs $(t_q, t_c)$ likely to exhibit a non-taxonomic relation. Second, each pair $(t_q, t_c)$ is converted into a natural language query $Q_{(t_q, t_c)}$ using a verbalization template, and the retriever selects top-$k$ relation candidates: $\mathcal{R}_k(t_q, t_c) = \operatorname{TopK}_{r \in \mathcal{R}} s(Q_{(t_q, t_c)}, r)$, where $s(\cdot)$ is the RAG similarity function. Finally, the LLM acts as a relation verifier with prediction function $LLM_r : \mathcal{T} \times \mathcal{T} \times \mathcal{R} \rightarrow [0,1]$, where $LLM_r(t_q, t_c, r) = 1$ indicates that relation $r$ holds. The final predicted relations are $\mathcal{R}^*(t_q, t_c) = \{ r \in \mathcal{R}_k(t_q, t_c) \mid LLM_r(t_q, t_c, r) = 1 \}$. Aggregating over all queries gives the discovered set:  
\[
\mathcal{E}_R^* =
\bigcup_{t_q \in \mathcal{T}}
\bigcup_{t_c \in \mathcal{T}_k(t_q)}
\{ (t_q, r, t_c) \mid r \in \mathcal{R}^*(t_q, t_c) \}.
\]
Following the OntoLearner design, this task reuses the taxonomy discovery and term typing formulations with minimal modifications. 


\section{Results}

\begin{figure*}[!htbp]
    \centering
    \small
    \begin{subfigure}{\textwidth}
        \centering
        \includegraphics[width=\textwidth]{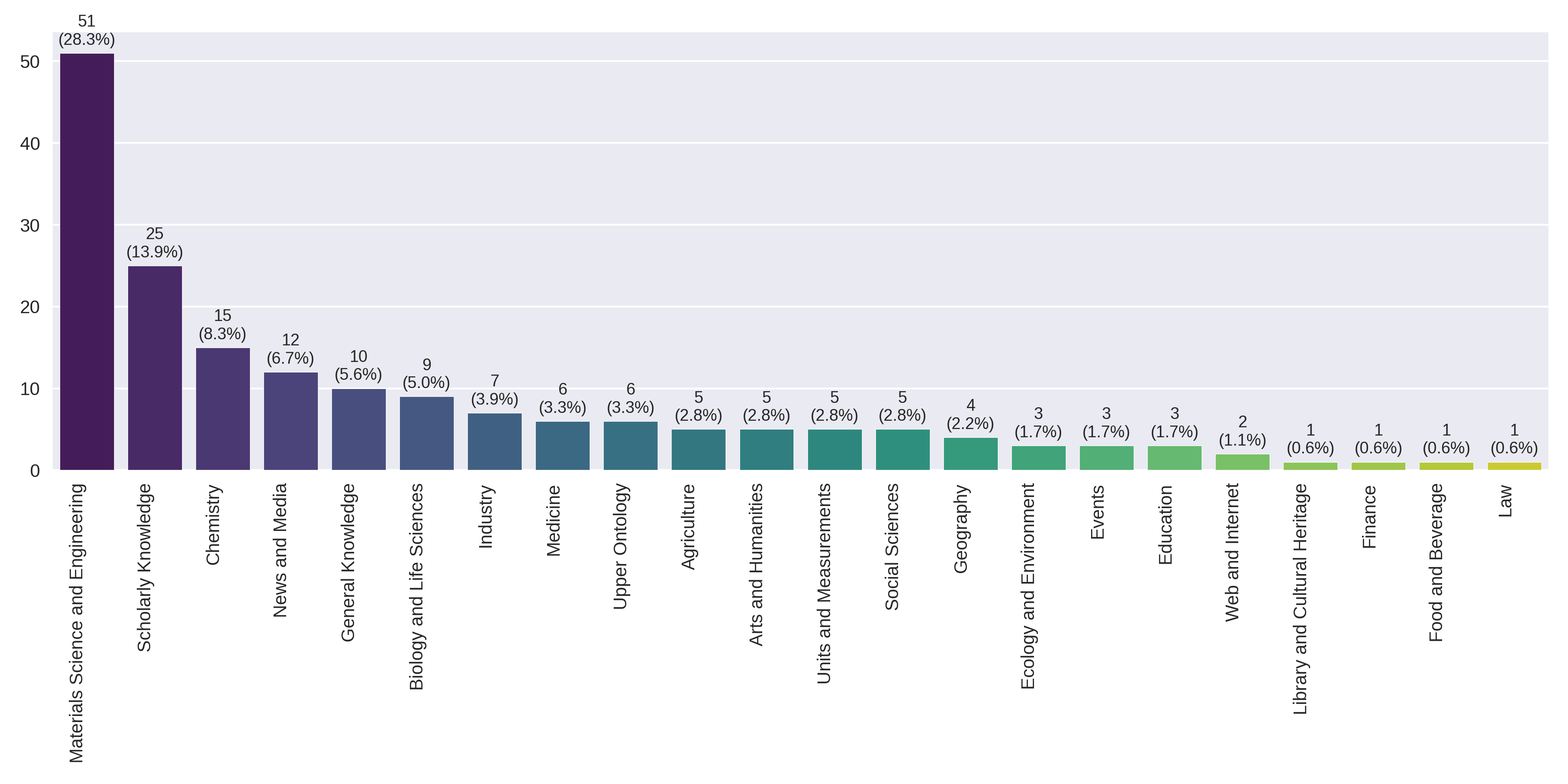}
        \caption{Distribution of 180 benchmark ontologies across 22 domains. The number of ontologies is mentioned above the bars with their percentage in comparison to the total ontologies in the collection.}
        \label{fig:domain-distribution}
    \end{subfigure}
    
    \begin{subfigure}{\textwidth}
        \caption{Summary statistics of key ontology metrics in OntoLearner, including mean, standard deviation (STD), quartiles (25\%, 50\%, 75\%), and maximum values.}
        \label{tab:benchmark-ontology-stats}
        \centering
        \small
        \begin{tabular}{l|r|r|r|r|r|r}
            \hline
            \textbf{Metric} & \textbf{Mean} & \textbf{STD} & \textbf{25\%} & \textbf{50\%} & \textbf{75\%} & \textbf{Max} \\
            \hline
            \hline
            \textit{Total Nodes} & 43,455	&260,522&	227	&659&	4,245	&2,433,610 \\
            \textit{Class No.}	&3,401	&18,672	&30&	99	&610	&220,816\\
            \textit{Properties No.}	&93	&266&	10&	35&	93	&3,029\\
            \textit{Individuals No.}	&7,654&	92,233	&0	&0	&15	&1,234,769\\
            \textit{Averaged Depth} &	2.27	&3.62	&0.57	&1.37	&2.71	&38.24\\
            \textit{Averaged Breadth} &	3,852	&27,830	&5	&16	&63	&310,545\\
            \textit{Processing Time (s)}	&25.12	&144.33	&0.03	&0.10	&0.95	&1196.53\\
            \hline
        \end{tabular}
    \end{subfigure}
    
    \begin{subfigure}{\textwidth}
        \centering
        
        \includegraphics[width=\textwidth]{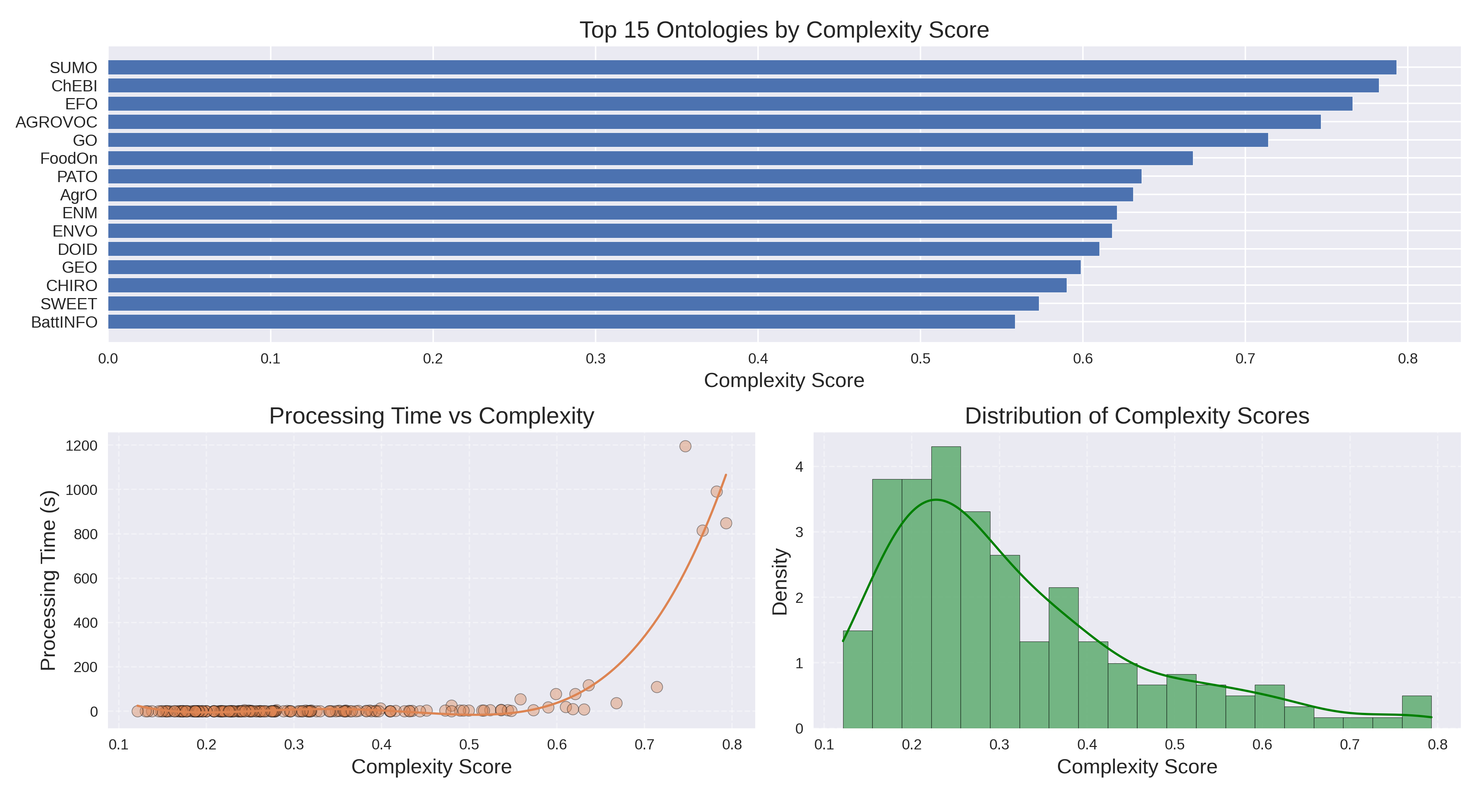}
        \caption{Structural Analysis: The top plot shows the top-15 ontologies' complexity scores, where the bottom-left plot shows the processing time vs. complexity score behaviour, and the bottom-right plot shows the distribution of complexity scores of ontologies.}        
        \label{fig:structural-analysis}
    \end{subfigure}
    \caption{Overview of the OntoLearner benchmark collection.}
    \label{fig:ontolearner-overview}

\end{figure*}

\subsection{Overview of Available Ontologies} 

The ontologies included in OntoLearner span diverse domains such as materials science and engineering, biology, life sciences, agriculture, law, library science, and cultural heritage. As of the date of this article submission, the collection comprises 180 ontologies containing 612,223 classes across 22 domains, and is designed to grow continuously. Ontologies were systematically collected using four objectives: 1) domain representativeness across major scientific disciplines, 2) structural diversity, 3) accessibility and licensing compatibility, and 4) active maintenance status. Each ontology was processed with OntoLearner's \textit{Ontologizer} component (described in \autoref{sec:software-core-components}) to ensure consistent metric computation and programmatic access. \autoref{fig:domain-distribution} depicts the distribution of ontologies. The \textit{Materials Science and Engineering}, with ontologies (28.3\%), is the most represented domain, followed by Scholarly Knowledge with 25 ontologies (13.9\%), while the 13 domains contain five or fewer ontologies each.

An ontology can be statistically presented via multiple metrics including graph metrics (i.e., nodes, edges), knowledge coverage metrics (i.e., classes, individuals, properties), hierarchical metrics (i.e., max, min, and mean depth), breadth metrics (i.e., max, min, and mean breadth), and dataset statistics (i.e., term types, taxonomic relations).
All metrics are automatically updated as new ontologies are added and are publicly accessible via a dedicated benchmark metrics space \footnote{\url{https://huggingface.co/spaces/SciKnowOrg/OntoLearner-Benchmark-Metrics}} and documentation website. A summary of these statistics is presented in \autoref{tab:benchmark-ontology-stats}, revealing substantial structural heterogeneity across the collection (high variability in STD). Most ontologies are small to medium in size, with the median substantially smaller than the mean, reflecting a right-skewed distribution. However, a small subset is extremely large, deep, and richly connected. This long-tail distribution reflects realistic ontology ecosystems, where lightweight vocabularies coexist with large, highly expressive knowledge models. 

To further capture ontological variability, OntoLearner introduces an ontological complexity score, $\mathcal{C}_{score}$ (described in \autoref{sec:ontology-complexity-scorer}), that summarizes structural characteristics into a single value. The structural analysis in \autoref{fig:structural-analysis} highlights how this score differentiates ontologies across the collection. Highly expressive ontologies such as SUMO~\cite{SUMO}, ChEBI~\cite{ChEBI}, EFO~\cite{EFO}, AGROVOC~\cite{AGROVOC}, and GO~\cite{GO1,GO2} appear among the most complex ($\mathcal{C}_{score} \approx 0.72$–$0.82$), reflecting their large class counts, deep hierarchies, and extensive branching structures (see top plot in \autoref{fig:structural-analysis}). The relationship between $\mathcal{C}_{score}$ and processing time (bottom-left plot) shows a clear nonlinear trend: ontologies with $\mathcal{C}_{score} < 0.5$ require minimal processing time, whereas computational cost increases sharply beyond $\mathcal{C}_{score} \approx 0.6$, with the most complex ontologies requiring up to 1,200 seconds to process. The distribution of $\mathcal{C}_{score}$ values across the collection (bottom-right plot) further confirms the long-tail pattern observed in the statistical summary, where most ontologies fall within moderate complexity ranges ($0.15$–$0.35$) and only a small subset are structurally and computationally demanding. Overall, this diversity in size, structure, and computational cost makes the OntoLearner collection particularly suitable for benchmarking OL methods, as it reflects the wide spectrum of ontology characteristics encountered in real-world knowledge engineering scenarios.

\subsection{Retrievers} 

\begin{figure*}[!htb]
    \centering
    \small
    \begin{subfigure}{\textwidth}
        \centering
        \includegraphics[width=\textwidth]{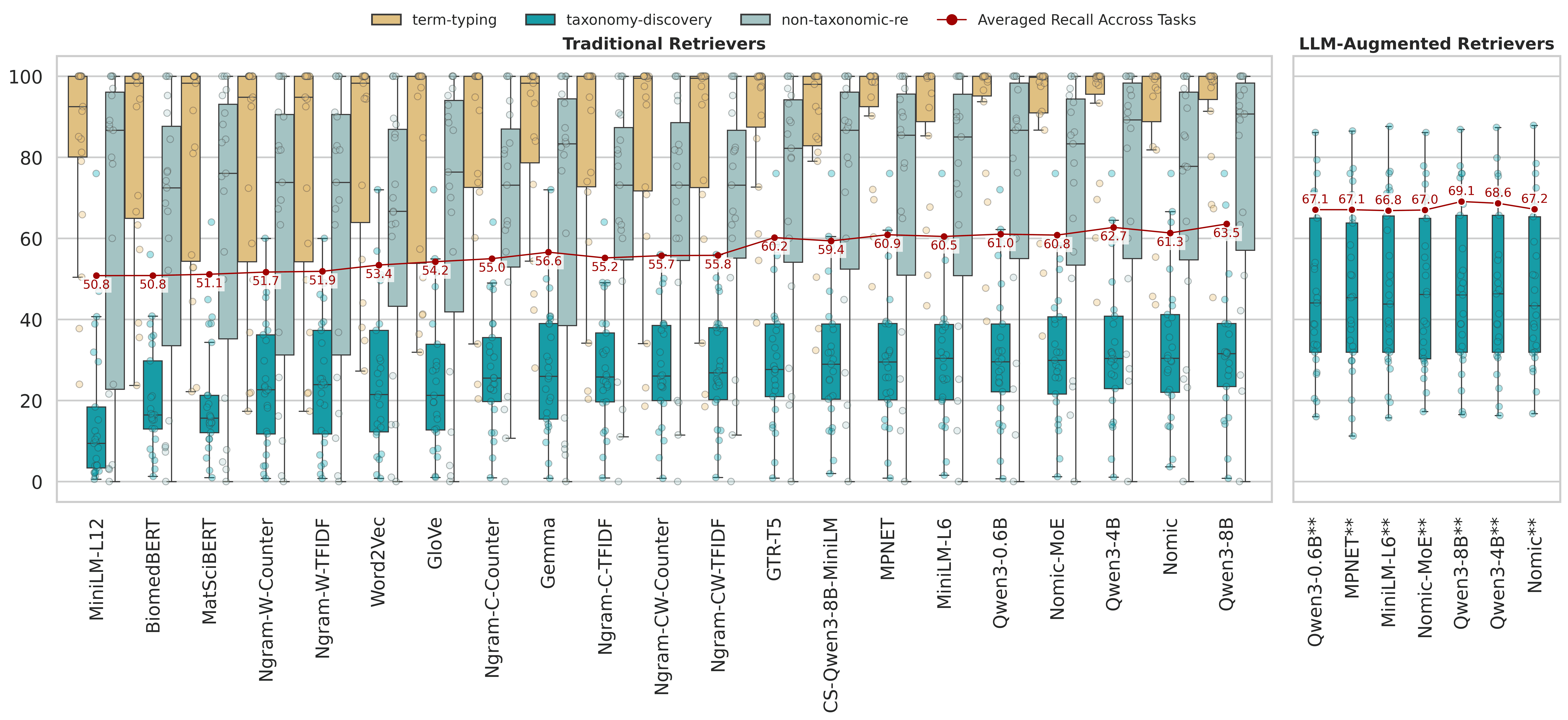}
        \caption{Recall performance of various retrieval models using top-k 15, by task. Boxplots summarize the distribution per model per task for our selected collection of 26 ontologies. The dots show individual recall per ontology. The models with the $^{**}$ symbol refer to the results of the augmented retriever approach. Retrievers on the x-axis are ordered by average recall of the type taxonomy discovery task. The red line chart represents each retriever's averaged recall across all three tasks. Augmented retrievers are only applied to the type taxonomy discovery task.}
        \label{fig:retrievers}
    \end{subfigure}
    
    \begin{subfigure}{\textwidth}
        \centering
        \includegraphics[width=\textwidth]{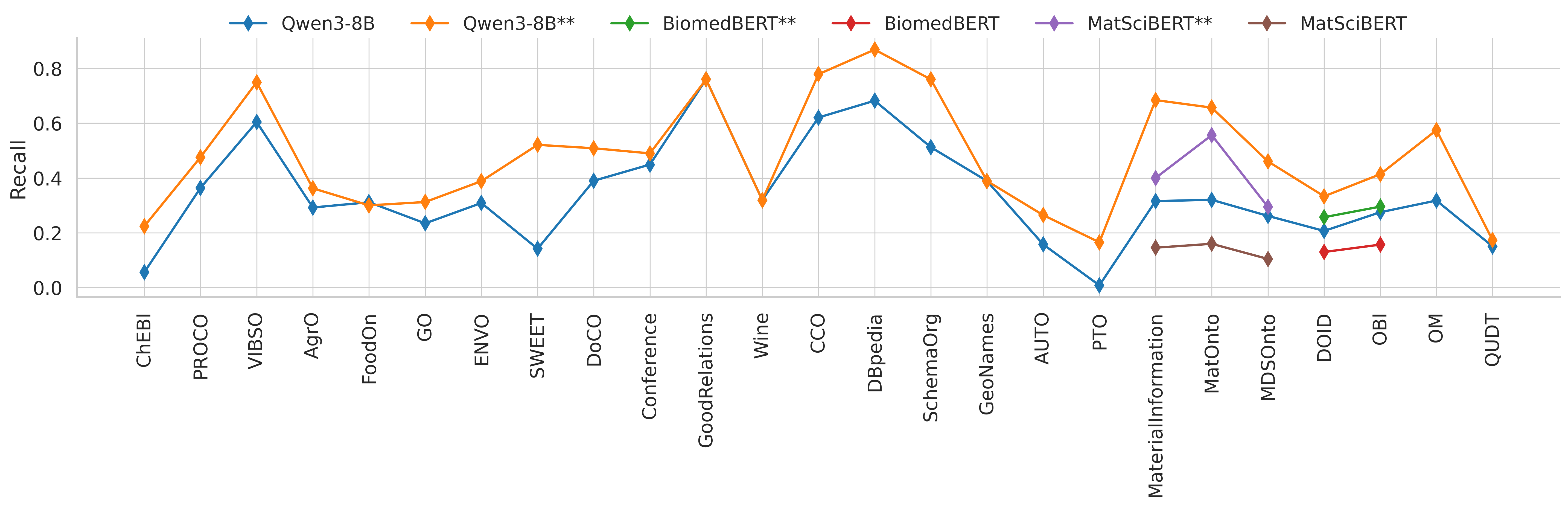}
        \caption{Recall comparison across ontologies for the taxonomy discovery task using the traditional \textit{Qwen3-8B} retriever and the LLM-augmented variant \textit{Qwen3-8B$^{**}$}. Domain-specific retrievers (\textit{BiomedBERT} and \textit{MatSciBERT}) are also shown where applicable. Results are reported using a top-$k$ value of 15. Ontologies are represented by their OntoLearner IDs on the x-axis; the corresponding full ontology names are listed in \autoref{tab:benchmark-ontologies}.}
        \label{fig:retriever-comparison}
    \end{subfigure}
    
    \caption{Retriever-based learners comparison.}
    \label{fig:retriever-results}
\end{figure*}

Effective OL depends on retrieving relevant ontological knowledge before any generation or classification. We evaluate 22 retrieval models spanning lexical baselines, classical embeddings, sentence transformers, dense retrievers, and domain-specialized models across the three OL tasks (learning tasks are detailed at \autoref{sec:software-core-components}), using \textit{Recall@15} as the primary metric. This broad comparison reveals where retrieval alone suffices, where it fails, and where LLM-Augmented retrieval helps. Results are reported on 26 ontologies selected to represent the structural diversity of the OntoLearner collection (see \autoref{sec:selected-retrievers-and-llms}).

\noindent\textbf{Task-Dependent Performance.} Results in \autoref{fig:retrievers} show clear task dependence. Term typing is generally easy, with most embedding models achieving high recall (median $>$95\%), with Qwen3 (0.6B, 4B, 8B) and Nomic variants averaging 89–92\%. Taxonomy discovery is much harder, with top models reaching only 9–37\% recall, reflecting the difficulty of capturing hierarchical relations through embeddings alone~\cite{yang2025achievinghyperboliclikeexpressivenessarbitrary,llms4ol2025overview,schmelzeisen2019learning}. Among classic retrievers, Qwen3-8B performs best for taxonomy discovery (averaged recall of $\approx$34\%). Non-taxonomic RE shows the greatest variability (near 0 to higher than 90\%), depending on domain and relation type, highlighting that some relations align well with embeddings (e.g., part-of, used-for) while others remain highly context-dependent. This gap confirms that retrieval alone cannot solve OL tasks.

\noindent\textbf{Impact of Model Architecture and Scale.} Lexical baselines (i.e, \textit{Ngrams}) achieve moderate recall on term typing (averaged recall of $\approx$80\%) and non-taxonomic relations ($\approx$64\%) but perform poorly on taxonomy discovery ($\approx$27\%), illustrating the limits of surface-form approaches. Traditional embeddings (GloVe, Word2Vec) show similar patterns ($\approx$81\%, 26\%, 63\%), though they suffer from out-of-vocabulary. Sentence transformers improve substantially ($\approx$90\%, 37\%, 70\%), with larger models like GTR-T5 and MPNET outperforming smaller MiniLM. Modern embeddings reach slightly higher recall ($\approx$91\%, 40\%, 71\%), with Qwen3-8B providing the best balance. However, scaling across 0.6B–8B yields diminishing returns. Nomic and Nomic-MoE perform similarly, demonstrating that bigger (8B) or more complex models (Nomic-MoE) alone cannot fully solve these tasks. Even though the simple embedding models work well for simple tasks, they hit a ceiling on taxonomy discovery.

\noindent\textbf{Domain-Specific Retrievers.} Domain-specialized models such as BiomedBERT~\cite{BiomedBERT} and MatSciBERT~\cite{MatSciBERT} perform poorly on general-domain ontologies, with an average recall of $\approx$51\%. Although pretrained on specialized corpora and strong on domain-focused NLP tasks, their retrieval results in OL indicate limited preservation of structured semantics. The \autoref{fig:retriever-comparison} further shows that LLM-based enhancement yields uneven improvements. While augmentation produces measurable gains, particularly within its respective domain, cross-domain generalization remains weaker than that of general-purpose retrievers. These findings suggest domain-specific pretraining biases the embedding space toward specialized terminology and local semantic patterns, hindering alignment with heterogeneous ontology structures. 

\noindent\textbf{Cross-Task Performance Consistency.} The red line in \autoref{fig:retrievers} shows averaged recall across all tasks, highlighting model robustness. Qwen3-8B and Nomic are the most consistent, maintaining high recall across diverse tasks. In contrast, specialized models exhibit task-specific variability: character-level n-grams excel at term typing but perform poorly on relational tasks. The hybrid CS-Qwen3-8B-MiniLM performs well on term typing and non-taxonomic relations but offers a marginal gain on taxonomy discovery over Qwen3-8B, suggesting that architectural complexity does not compensate for the task's difficulty. Nevertheless, no single architectural family consistently maintains strong performance across OL tasks. The gap between task-specific peaks and cross-task consistency reveals that single-method optimization systematically trades robustness for narrow performance gains. 

\noindent\textbf{Breaking the Retrieval Ceiling: LLM-Augmented Retrieval.} Applying LLM-Augmented retrieval to taxonomy discovery substantially boosts recall across top retrievers, improving average performance by 14–16\%. Notable gains include Nomic$^{**}$ (33.2\% $\rightarrow$ 47.9\%), Qwen3-4B$^{**}$ (32.9\% $\rightarrow$ 47.9\%), and Qwen3-8B$^{**}$ (33.8\% $\rightarrow$ 47.7\%). Mid-tier models benefit most: Nomic-MoE$^{**}$ (+15.6\%) and MiniLM-L6$^{**}$ (+16\%), narrowing the gap with top performers. Qwen3-8B$^{**}$ achieves the highest overall recall ($\approx$69.1\%, +5.6\% over base). By generating plausible parent candidates with an LLM before embedding-based ranking, this two-stage approach combines generative reasoning with discriminative retrieval, surpassing either component alone and exemplifying the hybrid infrastructure OntoLearner was designed to enable.

These results reveal a consistent pattern: \emph{no single retrieval architecture, regardless of scale, specialization, or complexity, can handle all OL tasks}. Term typing aligns with embeddings; taxonomy discovery does not; domain pretraining limits generalization; and scaling offers diminishing returns. These observations indicate that effective OL requires principled combinations of retrieval and generation. LLM-Augmented retrieval demonstrates that such hybrid approaches are both feasible and superior, and OntoLearner provides the infrastructure to evaluate and compare such approaches systematically.

\begin{table*}[!htb]
    \caption{Averaged F1-scores (\%) of end-to-end RAG pipeline using Qwen3-Embedding-8B as a retriever across multiple domains and tasks.  For each domain task, the F1-scores are averaged across all available ontologies. The number of ontologies per domain are: \textit{Events} (1), \textit{Finance} (1), \textit{Food \& Beverage} (1), \textit{Agriculture} (2), \textit{Geography} (1), \textit{Materials Science \& Engineering} (4), \textit{Biology \& Life Sciences} (1), \textit{Ecology \& Environment} (2), \textit{General Knowledge} (3), \textit{Industry} (2), Medicine (2), \textit{Units \& Measurements} (2), \textit{Chemistry} (3), and \textit{Education} (1). Best, second-best, and third-best results per task are highlighted using green ($\blacktriangle$), blue ($\bullet$), and red ($\blacklozenge$), respectively. The best-performing LLM per task is indicated in bold, while the best-performing LLM per domain is underlined. The ``\textit{I}'' refers to the instruction model, where the ``\textit{TH}'' refers to the thinking variant. 
    }
    \centering
    \label{tab:reranking-results}
    \resizebox{\textwidth}{!}{%
    \begin{tabular}{l| ccc| cc| c| cccc| c| c }
    \hline
     &
    \rotatebox{90}{\textbf{Qwen3-0.6B}} &
    \rotatebox{90}{\textbf{Qwen3-8B}} &
    \rotatebox{90}{\textbf{Qwen3-14B}} &
    \rotatebox{90}{\textbf{Qwen3-4B-I}} &
    \rotatebox{90}{\textbf{Qwen3-4B-TH}} &
    \rotatebox{90}{\textbf{Qwen3-Next-80B-I}} &
    \rotatebox{90}{\textbf{Gemma-3-1B-it}} &
    \rotatebox{90}{\textbf{Gemma-3-4B-it}} &
    \rotatebox{90}{\textbf{Gemma-3-12B-it}} &
    \rotatebox{90}{\textbf{Gemma-3-27B-it}} &
    \rotatebox{90}{\textbf{Falcon-H1-1.5B-Deep-I}} &
    \rotatebox{90}{\textbf{Mistral-Small-3.2-24B-I}} \\
    \hline
    \multicolumn{13}{l}{\textbf{\textit{Term Typing Task}}}\\
    \hline
         \textit{Events} & 50.0 & 56.2 & 40.0 & \underline{58.1} & 49.1 & 55.2 & \textbf{43.2} & 48.5 & 52.4 & 56.0 & 48.8 & 53.1 \\
         \textit{Finance} & 22.2 & 65.6 & \textbf{71.8} & 71.2 & 25.0 & \underline{\best{\textbf{82.7}}} & 21.4 & 65.7 & 62.3 & \textbf{64.2} & 54.5 & \textbf{68.3} \\
         \textit{Food \& Beverage} & 8.5 & 54.1 & 42.0 & 61.0 & 7.3 & \underline{61.3} & 11.9 & 29.3 & 30.8 & 49.5 & 47.3 & 42.7 \\
         \textit{Agriculture} & 39.5 & \underline{69.5} & 61.9 & 60.5 & 36.6 & 67.6 & 33.4 & 47.5 & 55.9 & 58.6 & 52.4 & 61.9 \\
         \textit{Geography} & \textbf{66.7} & \third{\textbf{81.9}} & 5.0 & \underline{\second{\textbf{82.3}}} & \textbf{56.6} & 57.2 & 12.1 & \textbf{72.1} & \textbf{78.7} & 27.5 & \textbf{65.5} & 47.5 \\
         \textit{Material Science \& Engineering} & 17.0 & 41.8 & 44.7 & 47.3 & 16.9 & \underline{63.7} & 14.0 & 21.1 & 29.3 & 44.6 & 36.0 & 33.0 \\
         \textit{Ecology \& Environment} & 14.6 & 47.4 & 49.1 & 47.8 & 13.7 & \underline{53.2} & 13.7 & 36.9 & 36.9 & 49.3 & 43.5 & 42.8 \\
         \textit{General Knowledge} & 12.8 & 61.7 & 70.0 & 61.6 & 14.6 & \underline{73.2} & 19.9 & 37.7 & 52.6 & 59.0 & 53.8 & 59.8 \\
         \textit{Industry} & 55.8 & 51.1 & 38.3 & 61.1 & 37.2 & 28.8 & 34.0 & 43.3 & 53.1 & \underline{62.2} & 49.0 & 38.1 \\
         \textit{Medicine} & 12.2 & 41.2 & 44.9 & \underline{49.0} & 11.3 & 59.1 & 9.7 & 20.8 & 29.7 & 32.8 & 34.8 & 25.6 \\
         \textit{Units \& Measurements} & 14.2 & 34.7 & 42.3 & 33.1 & 16.6 & \underline{44.6} & 13.5 & 26.3 & 29.3 & 30.8 & 26.5 & 23.3 \\
         \textit{Chemistry} & 39.6 & 44.4 & 37.7 & 50.0 & 29.0 & 46.0 & 22.9 & 38.8 & 43.1 & \underline{46.9} & 24.7 & 44.0 \\
        \hline
    \multicolumn{13}{l}{\textbf{\textit{Type Taxonomy Discovery Task}}}\\
        \hline
         \textit{Education} & 8.2 & 19.7 & 14.8 & 25.4 & 9.9 & 22.0 & 6.2 & 10.8 & 16.8 & 25.4 & \textbf{17.0} & \underline{25.6} \\
         \textit{Events} & 5.0 & 14.9 &  32.9 & 19.5 & 5.6 & \underline{\textbf{\second{34.4}}} & 6.3 & 6.5 & 13.3 & 27.5 & 9.6 & 29.1 \\
         \textit{Finance} & 8.4 & \textbf{20.0} & \underline{\textbf{\best{42.1}}} & 17.1 & 6.5 & \third{33.8} & \textbf{7.9} & 9.0 & 16.2 & 21.7 & 10.2 & 26.8 \\
         \textit{Food \& Beverage} & 7.4 & 18.6 & 14.7 & 24.0 & 8.8 & 20.0 & 7.0 & 8.1 & 14.3 & 17.7 & 9.3 & \underline{25.0} \\
         \textit{Agriculture} & 1.6 & 5.4 & \underline{8.5} & 5.1 & 1.7 & 8.4 & 1.8 & 2.1 & 4.1 & 5.7 & 3.1 & 5.9 \\
         \textit{Geography} & \textbf{11.0} & 19.7 & 19.0 & \textbf{25.5} & \textbf{12.5} & 24.2 & 4.8 & \textbf{14.1} & \textbf{26.1} & \underline{\textbf{31.6}} & 13.6 & \textbf{31.3} \\
         \textit{Material Science \& Engineering}  & 1.9 & 7.5 & \underline{15.8} & 7.1 & 2.2 & 14.4 & 2.0 & 2.6 & 7.3 & 9.9 & 3.1 & 11.2 \\
         \textit{Biology \& Life Science} & 2.0 & 6.2 & \underline{10.8} & 5.9 & 1.9 & 8.5 & 2.1 & 2.2 & 3.7 & 5.8 & 2.9 & 6.2 \\
         \textit{Ecology \& Environment} & 1.7 & 6.3 & \underline{15.5} & 7.1 & 1.9 & 12.9 & 2.0 & 2.3 & 5.4 & 8.4 & 2.4 & 8.7 \\
         \textit{General Knowledge} & 2.3 & 9.9 & \underline{26.7} & 12.3 & 3.0 & 21.2 & 2.6 & 4.0 & 10.7 & 14.6 & 4.3 & 17.5 \\
         \textit{Industry} & 1.7 & 9.4 & 12.3 & 9.9 & 2.1 & \underline{14.7} & 2.1 & 3.6 & 7.7 & 12.1 & 3.3 & 12.9 \\
         \textit{Medicine} & 2.1 & 9.3 & \underline{16.0} & 9.7 & 2.2 & 13.7 & 2.3 & 2.7 & 6.6 & 10.3 & 3.8 & 11.2 \\
         \textit{Units \& Measurements} & 3.0 & 10.5 & \underline{23.0} & 11.2 & 3.1 & 21.5 & 2.0 & 3.6 & 9.2 & 13.4 & 5.3 & 13.0 \\
         \textit{Chemistry} & 2.1 & 7.2 & \underline{13.5} & 6.9 & 2.1 & 12.2 & 2.2 & 2.4 & 6.4 & 7.9 & 3.3 & 8.4 \\
        \hline
    \multicolumn{13}{l}{\textbf{\textit{Non-Taxonomic Relation Extraction Task}}}\\
        \hline
         \textit{Education} & 7.0 & 11.6 & \underline{25.9} & 14.6 & 7.4 & 8.1 & 9.8 & 6.9 & 17.8 & 12.7 & 4.5 & 9.2 \\
         \textit{Events} & 0.0 & \textbf{40.0} & \textbf{42.9} & 26.7 & 26.7 & 37.5 & 46.2 & \underline{54.5} & \textbf{26.7} & 33.3 & 0.0 & 46.2 \\
         \textit{Finance} & 22.3 & 10.8 & 22.6 & 34.5 & 13.9 & \underline{36.3} & 23.6 & 21.6 & 23.9 & 30.1 & 14.0 & 31.5 \\
         \textit{Agriculture} & 0.2 & 0.4 & 0.4 & 0.4 & 0.2 & \underline{0.6} & 0.2 & 0.2 & 0.3 & 0.4 & 0.3 & 0.3 \\
         \textit{Material Science \& Engineering}  & 6.1 & 7.0 & \underline{12.2} & 10.5 & 4.6 & 11.6 & 6.1 & 5.5 & 10.3 & 9.4 & 3.1 & 10.1 \\
         \textit{Biology \& Life Science} & 12.4 & \underline{25.5} & 15.4 & 13.8 & 10.9 & 13.0 & 20.9 & 16.5 & 13.2 & 13.6 & 13.6 & 14.3 \\
         \textit{Ecology \& Environment} & 3.1 & 6.7 & \underline{9.0} & 7.1 & 2.9 & 9.5 & 2.9 & 3.7 & 6.9 & 5.8 & 2.4 & 7.5 \\
         \textit{General Knowledge} & 7.7 & 11.8 & \underline{12.2} & 11.0 & 9.2 & 9.0 & 6.8 & 6.0 & 11.5 & 8.7 & 7.4 & 8.5 \\
         \textit{Industry} & 19.3 & 23.0 & 13.1 & 17.1 & 17.3 & 20.8 & \underline{31.3} & 19.6 & 17.2 & 14.6 & 26.1 & 15.9 \\
         \textit{Medicine} & 3.0 & \underline{6.8} & 5.4 & 6.6 & 2.3 & 5.2 & 4.6 & 3.2 & 4.4 & 4.1 & 5.1 & 4.3 \\
         \textit{Units \& Measurements} & \textbf{\third{85.7}} & 0.0 & 15.4 & \textbf{58.8} & \textbf{\third{85.7}} & \textbf{\second{90.9}} & \textbf{50.0} & \textbf{66.7} & 15.4 & \textbf{50.0} & \underline{\textbf{\best{95.7}}} & \textbf{73.7} \\
         \textit{Chemistry} & 4.3 & \underline{13.8} & \underline{13.8} & 9.9 & 3.2 & 9.9 & 4.3 & 4.1 & 8.8 & 12.2 & 3.5 & 6.3 \\
        \hline
    \end{tabular}
    }
\end{table*}

\subsection{Reranking for the RAG Pipeline}
\autoref{tab:reranking-results} presents averaged F1-scores of the end-to-end RAG pipeline across domains, using \textit{Qwen3-Embedding-8B} as a retriever and 12 LLMs as rerankers. Unlike our initial \textsc{LLMs4OL} work~\cite{babaei2023llms4ol}, which constrained the search space, candidates come from the full ontology class space, simulating realistic conditions where models must identify correct relations among thousands of plausible alternatives~\cite{giglou2024llms4ol,giglou2025llms4ol}. This change shifts the difficulty landscape: non-taxonomic RE was hardest in our earlier work, but taxonomy discovery is now the dominant challenge, reflecting the complexity of hierarchical reasoning at scale.

\noindent\textbf{Task Difficulty Reflects Real-World Complexity.} A clear performance hierarchy emerges across tasks. Term typing achieves the highest F1-scores (40–83\%; e.g., Qwen3-Next-80B reaches 82.7\% in \textit{Finance}), taxonomy discovery performs poorly (single digits in \textit{Biology \& Life Science}, \textit{Agriculture}, and \textit{Ecology \& Environment}), and non-taxonomic RE falls in between, ranging from 0.2\% (\textit{Agriculture}) to 95.7\% (\textit{Units \& Measurements}, Falcon-H1-1.5B). Variability reflects task characteristics: in high-frequency domains like \textit{Units \& Measurements}, LLMs memorize relations, making retrieval mostly redundant, whereas sparse domains like \textit{Agriculture} benefit most from RAG, which bridges partial model knowledge and task requirements. Term typing is effectively a local classification problem, while taxonomy discovery requires inferring a global hierarchical structure. Neither retrieval nor LLM handles this reliably at scale, indicating that the retrieval ceiling is a fundamental property of the task.

\noindent\textbf{Scale Helps Term Typing But Not Taxonomy Discovery.} Scaling from Qwen3-0.6B to Qwen3-14B produces dramatic gains on term typing (\textit{Finance}: 22.2\% $\rightarrow$ 71.8\%, \textit{General Knowledge}: 12.8\% $\rightarrow$ 70.0\%) but negligible gains on taxonomy discovery (\textit{Biology \& Life Science}: 2.0\% $\rightarrow$ 10.8\%, \textit{Agriculture}: 1.6\% $\rightarrow$ 8.5\%). Scaling improves representation and instruction following, benefiting term typing and relation extraction. Taxonomy discovery, however, requires reasoning (transitive closure, subsumption, and lattice consistency) that current models may not fully encode.

\noindent\textbf{Output Discipline Matters More Than Reasoning Depth.} Qwen3-4B-Instruct consistently outperforms Qwen3-4B-Thinking across domains and tasks (\textit{Finance} term typing: 71.2\% vs 25.0\%; \textit{Education} taxonomy: 25.4\% vs 9.9\%). The thinking variant underperforms on structured extraction, which requires schema compliance, controlled formatting, and grounding in retrieved candidates. Alignment and output discipline matter more than raw reasoning depth for OL tasks.

\noindent\textbf{Domain Coverage Is the Binding Constraint for Hard Domains.}  \textit{Agriculture} and \textit{Biology \& Life Science} perform poorly on both taxonomy discovery and non-taxonomic RE, with taxonomy discovery in \textit{Biology \& Life Science} reaching only 2–11\% and non-taxonomic RE in \textit{Agriculture} averaging 0.2–0.6\%. Failures arise from limited domain coverage: specialized terminology (e.g., Latin species names, crop traits) and complex hierarchies (e.g, \texttt{GO} has 40,000+ terms) overwhelm embeddings and LLMs. In contrast, high-resource domains like \textit{Units \& Measurements} achieve 85–95\% F1 on non-taxonomic RE because relations are largely memorized, making retrieval less impactful. This asymmetry extends across domains: models that excel in high-resource domains do not transfer their performance to specialized ones. For example, Qwen3-14B scores 72\% in \textit{Finance} but <45\% in \textit{Medicine} and \textit{Units \& Measurements}, while Qwen3-8B scores 82\% in \textit{Geography} but 35–41\% in other domains. These results suggest that a single RAG pipeline may not generalize across the full breadth of OL domains.

\noindent\textbf{Model Family Matters Independently of Scale.} Gemma3-4B achieves 54.5\% on \textit{Events} non-taxonomic RE (the highest among all models) outperforming Qwen3-14B (42.9\%) and Qwen3-Next-80B (37.5\%). Gemma3-1B similarly exceeds much larger models on \textit{Biology \& Life Science} relation extraction (20.9\% vs Qwen3-14B's 15.4\%), and on \textit{Events} (46.2\% vs Qwen3-8B's 40.0\%). This pattern reflects Gemma's stronger calibration for relational phrase recognition~\cite{eberts2019span}, which is critical for span-based relation extraction, a capability that appears to arise from training alignment rather than parameter count. For non-taxonomic RE specifically, smaller Gemma models offer a substantial computational efficiency advantage over larger Qwen alternatives, underscoring that task-aware model selection outperforms defaulting to the largest available model.

\noindent\textbf{Memorization vs. Structural Alignment.} To verify whether strong LLM performance in resource-rich domains reflects memorization, we conducted perplexity-based contamination analysis~\cite{dong2024generalization} on 1,191 relation pairs across high-resource domains (i.e., \textit{Finance} and \textit{Units \& Measurements}) using the Qwen family of LLMs (details in supplementary materials section 3). \textit{Units \& Measurements} ontologies (OM~\cite{OM} and QUDT~\cite{QUDT}) show consistently low to moderate perplexity. In particular, OM exhibits low perplexity (mean: 128.7–251.6, with 29–61\% of samples below 100), providing strong evidence of pretraining exposure and thus a high likelihood of dataset contamination. While QUDT shows higher but decreasing perplexity with model scale (mean: 177.5–1133.6), indicating partial familiarity, but it is less consistently memorized than OM. In contrast, the \textit{Finance} domain (GoodRelations~\cite{GoodRelations}) exhibits extremely high perplexity (mean $>6,148$, with $\sim$93\% of samples exceeding 400) despite achieving 71.8–82.7\% F1 on term typing, demonstrating that strong performance can occur without direct memorization. This dissociation suggests that benchmark performance reflects structural task-domain compatibility rather than simple data memorization.

These results deepen the pattern established in the retriever analysis: \emph{performance in OL is determined by the alignment between task structure, domain characteristics, and model training—not by scale or architecture alone}. Taxonomy discovery resists improvements in both retrieval and generation because it requires reasoning that current models do not natively encode in the real-world search space. RAG adds the most value when models are partially informed, a pattern revealed only by benchmarking across diverse domains. Instruction tuning generally outperforms reasoning variants because structured extraction demands output discipline, not open-ended inference. Domain coverage constrains the hardest tasks more than model capability does.

\subsection{Analysis}

To understand OL task behavior beyond aggregate metrics, we analyze two ontologies with contrasting scale and structure: \textit{Gene Ontology (GO)}\footnote{\url{https://ontolearner.readthedocs.io/benchmarking/biology\_and\_life\_sciences/go.html}}~\cite{GO1,GO2}, a large hierarchical ontology with $\mathcal{C}{score}=0.71$, and \textit{Materials and Data Science (MDSOnto)}\footnote{\url{https://ontolearner.readthedocs.io/benchmarking/materials\_science\_and\_engineering/mdsonto.html}}~\cite{MDSOnto}, a mid-sized ontology with $\mathcal{C}{score}=0.38$. These provide a controlled setting to examine whether failures arise from model limitations or ontological structure.

\begin{figure}[!htbp]
    \centering

    \begin{subfigure}{0.5\textwidth}
        \centering
        \includegraphics[width=\linewidth, height=10cm]{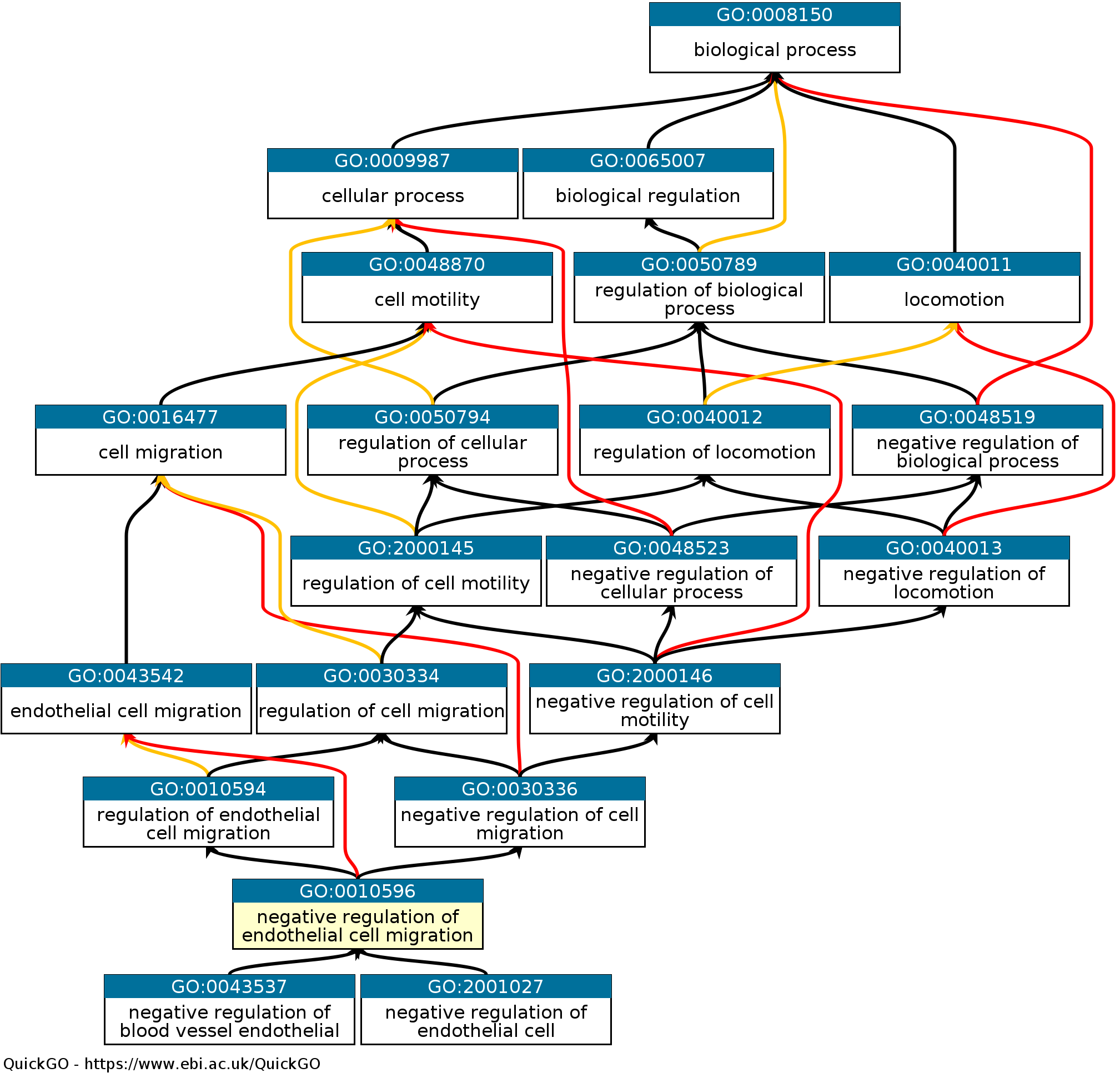}
        \caption{Gene Ontology is a large hierarchical ontology with 62,046 classes, 156,430 taxonomic relations. This char presents ancestor term GO:0010596 (\emph{negative regulation of endothelial cell migration}). \textcolor{black}{\textbf{Black}} arrows indicate \emph{is--a} relationships, \textcolor{red}{\textbf{Red}} arrows denote the non-taxonomic relationship \emph{negatively regulates}, and \textcolor{orange}{\textbf{Orange}} arrows denote the non-taxonomic relationship \emph{regulates}. The diagram was obtained from QuickGO (specifically from \url{https://www.ebi.ac.uk/QuickGO/term/GO:0010596}).}
        \label{fig:gen-ontology-subset}
    \end{subfigure}

    \begin{subfigure}{0.5\textwidth}
        \centering
        \includegraphics[width=\linewidth, height=9cm]{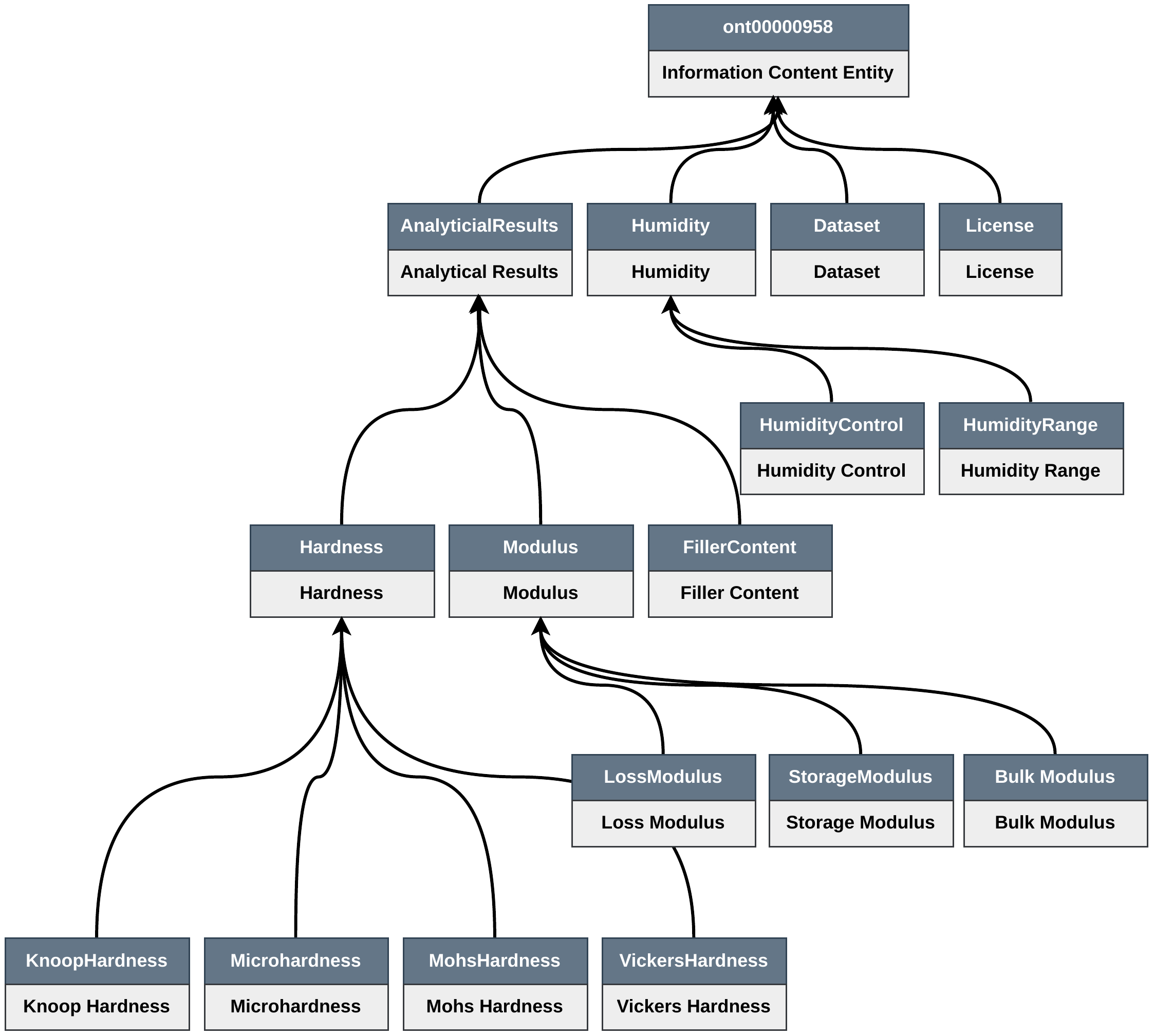}
        \caption{Ancestor chart for the Material Data Science Ontology. \textcolor{black}{\textbf{Black}} arrows indicate \emph{is--a} relationships.}
        \label{fig:mdsonto-ontology-subset}
    \end{subfigure}
    
    \caption{Illustrations of smoke tests.}
    \label{fig:smoke-tests-vis}
\end{figure}

\begin{figure*}[!htb]

\begin{subfigure}{\textwidth}
    \centering
    \footnotesize
    \caption{Recall@15 for different retrieval models across taxonomy discovery and non-taxonomic RE tasks for the GO and MDSOnto.}
    \label{tab:go-mdsonto-retrieval-results}
    \begin{tabular}{l|r|r|r|r|r|r|r|r|r|r|r|r|r|r}
        \hline
         & \rotatebox{90}{\textbf{Qwen3-8B**}} & \rotatebox{90}{\textbf{Qwen3-4B**}} & \rotatebox{90}{\textbf{Qwen3-0.6B**}} & \rotatebox{90}{\textbf{Qwen3-8B}} & \rotatebox{90}{\textbf{MPNET**}} & \rotatebox{90}{\textbf{Qwen3-4B}} & \rotatebox{90}{\textbf{MiniLM-L6**}} & \rotatebox{90}{\textbf{Qwen3-0.6B}} & \rotatebox{90}{\textbf{Nomic-MoE**}} & \rotatebox{90}{\textbf{MPNET}} & \rotatebox{90}{\textbf{MiniLM-L6}} & \rotatebox{90}{\textbf{Nomic**}} & \rotatebox{90}{\textbf{Nomic-MoE}} & \rotatebox{90}{\textbf{Nomic}} \\
        \hline
        
        \multicolumn{15}{l}{\textbf{Gene Ontology (GO)}}\\
        \hline
            \textbf{Taxonomy Discovery} & 31 & 31 & 30 & 23 & 30 & 23 & 30 & 22 & 30 & 20 & 20 & 31 & 22 & 23 \\
            \textbf{Non-Taxonomic RE} & 97 & 97 & 97 & 97 & 90 & 97 & 90 & 97 & 83 & 90 & 90 & 77 & 83 & 77 \\
            \textbf{Average} & 64 & 64 & 63 & 60 & 60 & 60 & 60 & 59 & 57 & 55 & 55 & 54 & 52 & 50 \\
        \hline
        \multicolumn{15}{l}{\textbf{Material Data Science Ontology (MDSOnto)}}\\
        \hline
            \textbf{Taxonomy Discovery} & 46 & 47 & 47 & 26 & 45 & 26 & 44 & 24 & 46 & 22 & 21 & 43 & 26 & 22 \\
            \textbf{Non-Taxonomic RE} & 42 & 25 & 12 & 42 & 13 & 25 & 13 & 12 & 16 & 13 & 13 & 28 & 16 & 28 \\
            \textbf{Average} & 44 & 36 & 29 & 34 & 29 & 25 & 28 & 18 & 31 & 17 & 17 & 35 & 21 & 25 \\
        \hline
    \end{tabular}
\end{subfigure}

    \vspace{0.8em}
    
\begin{subfigure}{\textwidth}
     \centering
     \includegraphics[width=\textwidth]{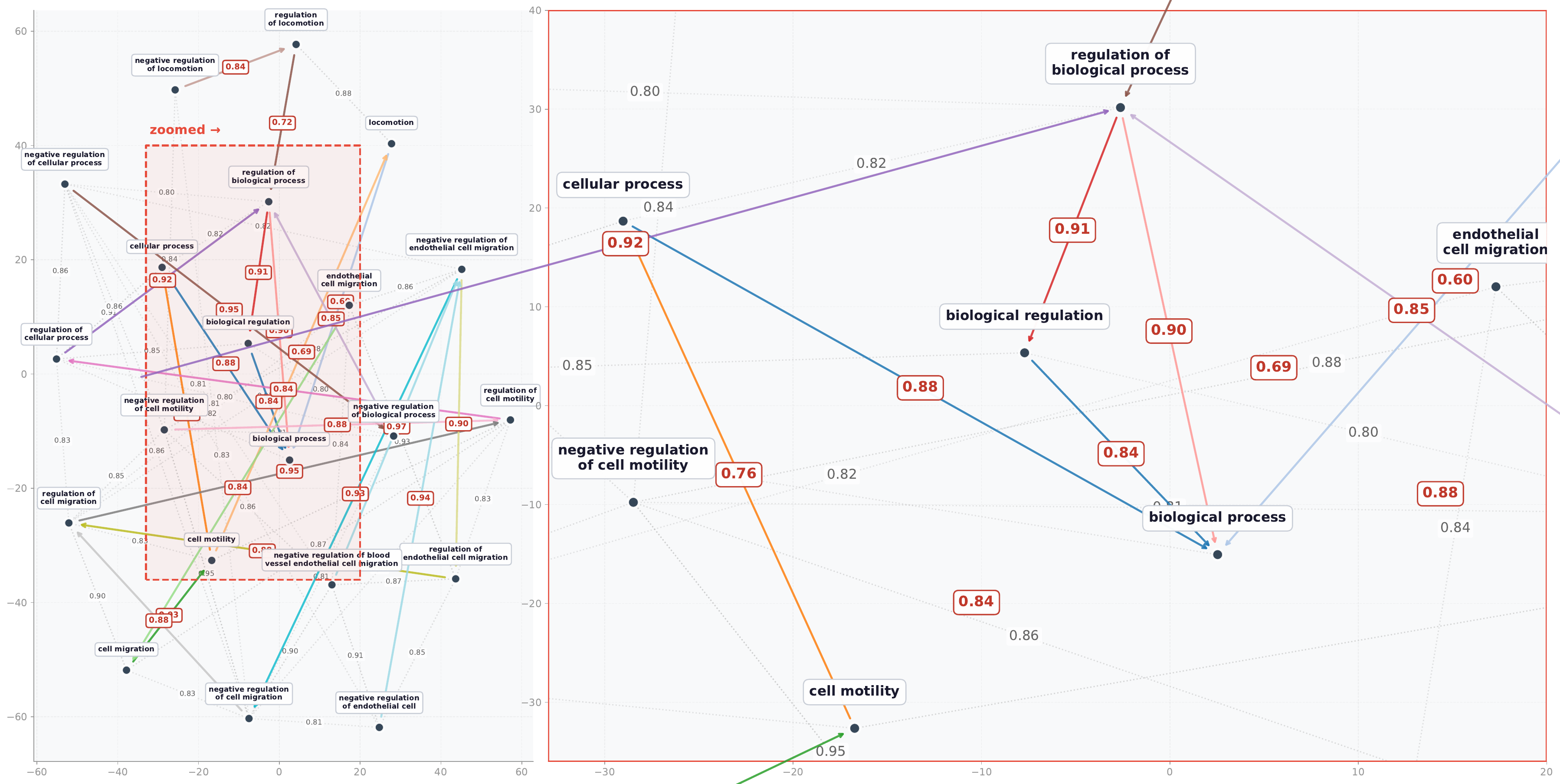}
     \caption{t-SNE visualization of selected smoke-test taxonomic pairs from GO, generated using the Qwen3-Embedding-8B model. The plot illustrates crowded clustering in the embedding space, which contributes to lower performance in the RAG retrieval module. Colored lines indicate taxonomic relations, while dotted lines represent incorrect relations with similarity of higher than 80. The red box highlights a dense region shown as a zoomed view.}
     \label{fig:tsne-visulaiation-smoke-test}
\end{subfigure}

    \caption{Retriever Limitations and Failure Modes.}
    \label{fig:retriever-limitations}
    
\end{figure*}

\begin{figure*}[!htb]

\begin{subfigure}{\textwidth}
    \centering
    \includegraphics[width=\linewidth]{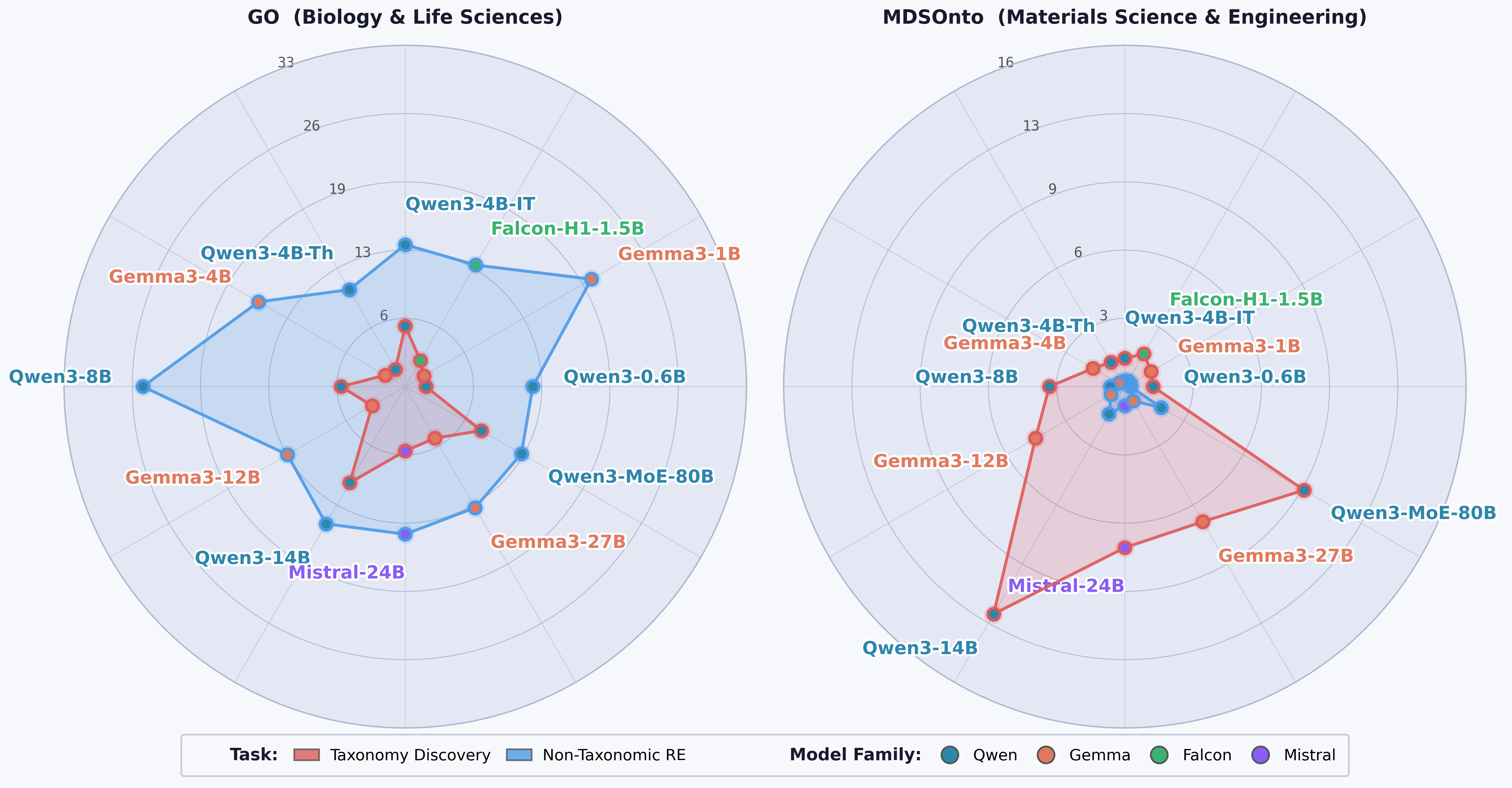}
    \caption{F1 scores of LLMs on taxonomy discovery and non-taxonomic RE across GO and MDSOnto ontologies.}
    \label{fig:go-vs-mdsonto-radar}
\end{subfigure}

\begin{subfigure}{\textwidth}
\centering
\caption{Model performance on GO and MDSOnto ontologies. Accuracies shown as percentages for 393 samples from GO -- 260 for taxonomy discovery + 133 for non-taxonomic re tasks -- and 223 samples from taxonomy discovery tasks using MDSOnto. Entropy reported as mean$\pm$std. $\Delta_{\text{calib}}$ = entropy difference (correct $-$ incorrect); positive values indicate good calibration. FP/FN = False Positive/Negative counts. Hall = hallucinated samples.}
\label{tab:case-studies-behavior-analysis}
\resizebox{\textwidth}{!}{%
\begin{tabular}{l|ccc|cc|cc|cccc|cc}
\hline
\multirow{2}{*}{\textbf{Model}} & \multicolumn{3}{c|}{\textbf{F1-Score (\%)}} & \multicolumn{2}{c|}{\textbf{Attention}} & \multicolumn{2}{c|}{\textbf{Calibration}} & \multicolumn{4}{c|}{\textbf{Error Bias}} & \multicolumn{2}{c }{\textbf{Error Types}} \\

& GO$_T$ & GO$_R$ & MDS$_T$ & GO$_{entropy}$ & MDS$_{entropy}$ & $GO_{\Delta_{\text{calib}}}$ & $MDS_{\Delta_{\text{calib}}}$ & GO$_{FP}$ & GO$_{FN}$ & MDS$_{FP}$ & MDS$_{FN}$ & GO$_{Hall}$  & MDS$_{Hall}$ \\
\hline
Qwen3-0.6B & 41.9 & 16.7 & 39.7 & 0.67$\pm$0.03 & 0.69$\pm$0.01 & 0.024 &0.005 & 201 & 0 & 167 & 0 & 108 & 75 \\
Gemma-3-1b-it & 0.0 & 10.5 & 3.8 & 2.56$\pm$0.14 & 2.63$\pm$0.02 & -0.163 &0.014 & 12 & 75 & 0 & 51 & 8 & 0 \\
Falcon-H1-1.5B-Deep & 64.7 & 13.2 & 27.9 & 1.75$\pm$0.08 & 1.80$\pm$0.03 & -0.110 &-1.805 & 127 & 10 & 14 & 43 & 97 & 3 \\
Gemma-3-4b-it & 44.1 & 17.3 & 52.3 & 1.78$\pm$0.08 & 1.82$\pm$0.01 & -0.026 &0.006 & 238 & 4 & 58 & 15 & 142 & 16 \\
Qwen3-4B-Instruct-2507 & 73.0 & 18.5 & 52.3 & 1.12$\pm$0.07 & 1.16$\pm$0.01 & 0.072 &0.002 & 141 & 4 & 10 & 32 & 116 & 2 \\
Qwen3-4B-Thinking-2507 & 0.0 & 0.0 & 0.0 & 1.09$\pm$0.05 & 1.10$\pm$0.01 & -1.088 &-1.098 & 0 & 1 & 0 & 3 & 0 & 0 \\
Qwen3-8B & 73.0 & 16.1 & 51.2 & 1.53$\pm$0.08 & 1.58$\pm$0.01 & 0.101 &0.003 & 159 & 4 & 6 & 34 & 139 & 1 \\
Gemma-3-12b-it & 61.9 & 16.2 & 35.0 & 1.37$\pm$0.11 & 1.42$\pm$0.01 & 0.117 &0.001 & 199 & 0 & 11 & 41 & 158 & 0 \\
Qwen3-14B & 78.4 & 17.2 & 22.2 & 1.44$\pm$0.08 & 1.48$\pm$0.01 & 0.056 &0.002 & 130 & 9 & 2 & 47 & 116 & 0 \\
Mistral-Small-3.2-24B & 75.1 & 15.9 & 49.4 & 1.83$\pm$0.11 & 1.86$\pm$0.01 & 0.152 &0.008 & 155 & 4 & 3 & 36 & 132 & 0 \\
Gemma-3-27b-it & 71.9 & 15.5 & 51.8 & 0.98$\pm$0.05 & 0.99$\pm$0.01 & 0.065 &-0.000 & 174 & 0 & 5 & 34 & 146 & 0 \\
Qwen3-Next-80B-A3B & 73.9 & 16.2 & 34.3 & 2.71$\pm$0.11 & 2.77$\pm$0.01 & 0.143 &0.003 & 156 & 4 & 3 & 43 & 132 & 0 \\
\hline
\end{tabular}%
}
\end{subfigure}
    \caption{LLM Behavior Analysis.}
    \label{fig:llm-behavior-analysis}
\end{figure*}

\noindent\textbf{Retriever Limitations and Failure Modes}. \autoref{tab:go-mdsonto-retrieval-results} shows \textit{Recall@15} across retrieval models for both ontologies. Despite improvements from LLM-Augmented variants, recall on taxonomy discovery remains low for both ontologies (see \autoref{tab:go-mdsonto-retrieval-results}), motivating a closer examination of why embeddings fail to separate hierarchical relations. To understand why embeddings fail to separate taxonomic relations clearly, we designed a smoke test on a carefully selected GO subset forming a multi-level hierarchy across several depth levels (see \autoref{fig:gen-ontology-subset}), and visualized the embedding space using t-SNE with \textit{Qwen3-Embedding-8B} (see \autoref{fig:tsne-visulaiation-smoke-test}). Three failure modes emerge consistently.

First, embedding representations collapse under dense compositional labels. Concepts such as ``regulation of cell migration'', ``cell migration'', and ``regulation of cell motility'' occupy nearly identical regions despite representing distinct hierarchy levels. This collapse is particularly pronounced in GO, where combination labels (e.g., ``regulation'', ``negative regulation'', ``cell'',``migration'') produce semantically saturated representations with minimal vector encodings. Second, similarity is distorted by hubness~\cite{nielsen2024hubness}. High-similarity links ($>$0.8) appear between unrelated concepts (dotted lines in \autoref{fig:tsne-visulaiation-smoke-test}), while genuine parent–child pairs span a wide overlapping range (0.60–0.97). For instance, ``negative regulation of endothelial cell migration'' shows 0.86 similarity to ``regulation of locomotion'' despite no taxonomic relation. This overlap makes threshold-based retrieval inherently unreliable. Third, embeddings saturate at a semantic ceiling where relatedness is preserved, but taxonomic entailment is lost. The model captures that ``cell migration'', ``cell motility'', and ``locomotion'' are related, but cannot encode the directional specificity required for \textit{is-a}. Structurally distinct pairs such as [``biological process'', ``regulation of biological process''] with similarity 0.91 and [``cellular process'', ``regulation of cellular process''] with similarity 0.88 produce near-identical similarity patterns.

Collectively, these behaviors point to a core technical limitation of embedding anisotropy~\cite{machina-mercer-2024-anisotropy,razzhigaev2024shape}, where representations occupy a narrow cone in vector space, causing cosine similarities to cluster in a poorly discriminative range for \textit{is-a} relations. Critically, the severity of this effect is determined by the ontology properties (i.e., label compositionality, class density, and relational complexity) rather than model scale. Embeddings encode relatedness, not entailment, and scaling does not resolve this mismatch when the ontology amplifies it.

\noindent\textbf{LLM Behavior Analysis}. \autoref{tab:case-studies-behavior-analysis} and \autoref{fig:go-vs-mdsonto-radar} reveal three behavioral patterns through attention, calibration, error bias, and hallucination analysis. \textit{1) Task asymmetry.} In controlled smoke tests, models achieve high F1 in taxonomy discovery ($\approx$70–75\%) but collapse to $\approx$15–18\% in non-taxonomic RE (e.g., Qwen3-8B: 73.0\% vs 16.1\%; Mistral-24B: 75.1\% vs 15.9\%). This exposes a fundamental weakness in non-hierarchical reasoning. Qwen3-4B-Thinking reaches 0.0\% F1, showing that open-ended chain-of-thought reasoning conflicts with the strict schema compliance required for structured OL extraction. Reasoning depth does not substitute for output discipline. \textit{2) Scaling helps selectively.} Despite a $\approx4\times$ increase in entropy from smaller to larger models (Qwen3-0.6B: 0.67 $\rightarrow$ Qwen3-Next-80B: 2.71), only marginal improvements appear in relational reasoning (Gemma-3-1B: 10.5\% $\rightarrow$ Gemma-3-27B: 15.5\% on GO non-taxonomic RE). Therefore, entropy reflects the sample's structural complexity rather than the model's capability, and it is not a reliable predictor of task performance. \textit{3) Model behavior diverges by ontology.} Calibration shifts dramatically across ontologies for identical models(e.g., $\Delta_{\text{calib}}=-1.088$ and $-1.805$), indicating that even when models predict incorrectly, predictions are made with high confidence. Calibration is therefore not an intrinsic model property but a \textit{model $\times$ domain} interaction. 

Error bias analysis reveals a structural effect. The FP/FN ratio reverses between ontologies, with systematic over-prediction on GO (Qwen3-0.6B: 201 FP vs 0 FN; Gemma-12B: 199 FP vs 0 FN; Qwen3-8B: 159 FP vs 4 FN), and under-prediction on MDSOnto (Gemma-1B: 0 FP vs 51 FN; Qwen3-Next-80B: 3 FP vs 43 FN). Hallucination follows the same pattern; GO produces up to 158 hallucinations per model while MDSOnto produces 0–3. These reversals correlate with ontological properties such as class distribution, relational density, and hierarchy, not architecture or scale.

These results converge on a single structural conclusion: \emph{in OL, failure modes are ontology-determined rather than model-determined}. Embedding anisotropy, calibration collapse, error bias reversal, and hallucination concentration scale with ontological complexity rather than model size.

\section{Discussion}
That failure in OL is ontology-determined rather than model-determined, mirroring the fragmentation identified across three decades of OL research, in which each generation optimized for narrow settings while the broader challenge remained unsolved. Therefore, the path forward requires not just better models but also ontology-aware architectures, richer evaluation frameworks, and infrastructure capable of systematically exposing these structural effects.

Overcoming the retrieval ceiling requires moving beyond single-architecture solutions. The smoke test analysis points to three concrete directions. First, contrastive learning objectives~\cite{gao2021simcse} designed to separate entailment from relatedness would directly address embedding collapse and saturation. Second, hyperbolic embedding spaces~\cite{sinha2024learning} are well-suited to hierarchical structures and could alleviate the hubness effect by explicitly representing taxonomic depth. Third, anisotropy mitigation techniques, such as whitening transformations, address the geometric crowding that makes cosine similarity unreliable for \textit{is-a}- typed pairs. Hybrid architectures that combine these approaches with symbolic features represent the most promising near-term direction for taxonomy discovery.

Current OL evaluation practice relies almost exclusively on F1-score, which is insufficient to characterize the failure modes this benchmark exposes. Four findings demand broader evaluation criteria. First, contamination analysis demonstrates that strong performance can occur both with and without pretraining exposure: \textit{Finance} achieves high F1 (71.8–82.7\%) despite extreme perplexity (mean 6,148), while \textit{Units \& Measurements} achieves similar F1 with low perplexity (mean 128.7–251.6), confirming that benchmark performance reflects structural task-domain compatibility rather than simply data leakage—a distinction F1 alone cannot reveal. Second, the error-bias reversal (extreme overprediction on GO, systematic underprediction on MDSOnto) is invisible to F1 but reveals that the model's behavior is governed by ontological structure rather than learned relational understanding. Third, the calibration analysis shows that models make confident predictions regardless of correctness and that calibration is a \textit{model $\times$ domain} interaction rather than an intrinsic model property, with the distinction only visible through cross-ontology evaluation. Fourth, the output discipline finding establishes that instruction-tuned models systematically outperform reasoning-optimized variants on structured extraction, suggesting that model selection for OL workflows should prioritize alignment and formatting discipline over raw reasoning capability. A single RAG pipeline optimized for high-resource domains will not generalize to specialized ones. Domain-specific retriever fine-tuning or ontology-aware pipeline design is required for consistently strong performance across the full breadth of OL.

The convergence of ontologies across 22 domains, three tasks, and more than 30 models points to a conclusion the field has approached but never been able to state precisely: the bottleneck in OL is not model capability but the structural mismatch between how models encode knowledge and how ontologies organize it. Closing this gap requires infrastructure that makes the mismatch visible, reproducible, and comparable across methods; precisely what has been missing.  OntoLearner provides that infrastructure, 180 ontologies spanning the full complexity spectrum, standardized evaluation across retrieval and LLMs, and the capacity to evaluate and compare hybrid approaches reproducibly, making findings from approaches like LLM-Augmented retrieval reproducible rather than observational. Our commitment to advancing this agenda is reflected in the LLMs4OL challenges at ISWC~\cite{giglou2024llms4ol,llms4ol2025overview}, which establish shared tasks and community benchmarks for reproducible OL research. More broadly, OntoLearner aims to support a new generation of hybrid knowledge-engineering workflows in which human expertise, curated ontologies, and foundation models collaborate—not sequentially, but as components of a unified, evaluable system.

\section{Conclusion}
OntoLearner is a modular, extensible Python library that addresses a key gap in Semantic Web tooling by integrating LLMs into OL workflows. It unifies curated ontologies, machine-readable formats, and reusable pipelines—enabling benchmarking and development of LLM-based methods for tasks like term typing, taxonomy induction, and relation discovery. Its design promotes reproducibility, FAIR compliance, and community contributions, positioning it as a foundation for advancing neuro-symbolic ontology engineering.   As the ecosystem of foundation generative AI models grows, OntoLearner is well-positioned to become a central resource for bridging formal knowledge representations with the generative capabilities of modern AI, advancing both methodological innovation and practical adoption in the Semantic Web.\\ \\

\noindent\textbf{Data and Code availability.} The OntoLearner benchmark collections are available at \href{https://huggingface.co/collections/SciKnowOrg/ontolearner-benchmarking-6823bcd051300c210b7ef68a}{https://huggingface.co/collections/SciKnowOrg/ontolearner-benchmarking}. Source code for OntoLearner is available under the MIT License at \url{https://github.com/sciknoworg/OntoLearner/}, with documentation at \url{https://ontolearner.readthedocs.io/}. \\ 

\noindent\textbf{Acknowledgements.} This work is jointly supported by the \href{https://scinext-project.github.io/}{SCINEXT project} (BMFTR, German Federal Ministry of Research, Technology and Space, Grant ID: 01lS22070), the KISSKI AI Service Center (BMFTR, Grant ID: 01IS22093C), the \href{https://www.nfdi4datascience.de/}{NFDI4DataScience initiative} (DFG, German Research Foundation, Grant ID: 460234259), and the \href{https://nfdi4ing.de/}{NFDI4ING initiatives} (DFG, German Research Foundation, Grant ID: 442146713).

\bibliographystyle{aaai}
\bibliography{main}

\end{document}